\documentclass[accepted]{uai2022_preprint} % for initial submission
\usepackage[american]{babel}
\usepackage{natbib} % has a nice set of citation styles and commands
\usepackage{mathtools} % amsmath with fixes and additions
\usepackage{booktabs} % commands to create good-looking tables
\usepackage{tikz} % nice language for creating drawings and diagrams

\usepackage{algorithm}
\usepackage{algpseudocode}
\usepackage{amsmath, amssymb, amsthm}

\DeclareMathOperator{\argmin}{arg\,min}
\DeclareMathOperator{\argmax}{arg\,max}
\newcommand{\R}{\mathbb{R}}
\newcommand{\E}{\mathbf{E}}

\title{Gradient-Matching Coresets for Rehearsal-Based Continual Learning}

\author[1]{\href{mailto:<balleslb@amazon.com>?Subject=Your UAI 2022 paper}{Lukas Balles}}
\author[1]{Giovanni Zappella}
\author[1]{Cédric Archambeau}
\affil[1]{%
    Amazon Web Services, Berlin
}

\begin{document}
\maketitle

\begin{abstract}
  The goal of continual learning (CL) is to efficiently update a machine learning
  model with new data without forgetting previously-learned knowledge.
  Most widely-used CL methods rely on a \emph{rehearsal memory} of
  data points to be reused while training on new data.
  Curating such a rehearsal memory to maintain a small, informative
  subset of all the data seen so far is crucial to the success of these methods.
  We devise a coreset selection method for rehearsal-based continual learning.
  Our method is based on the idea of gradient matching:
  The gradients induced by the coreset should match, as closely as possible,
  those induced by the original training dataset.
  Inspired by the neural tangent kernel theory, we perform this gradient matching
  across the model's initialization distribution, allowing us to extract a coreset
  without having to train the model first.
  We evaluate the method on a wide range of continual learning scenarios and
  demonstrate that it improves the performance of rehearsal-based CL methods compared
  to competing memory management strategies such as reservoir sampling.

  % Several industrial applications require machine learning models to be retrained
  % over time as new data become available. This is not a simple task since not
  % all the data observed over time can be stored or re-training from scratch is too costly.
  % These problems sparked the interest of the research community that in the last
  % years developed a number of solutions for the so-called Continual Learning.
  % Among these techniques, an important family of algorithms is the reharshal-based.
  % These algorithms usually rely on a small memory in which they store a portion of
  % the data to be used at later stage during the re-training. Curating the rehearsal memory
  % is a crucial task for obtaining good predictive performance.
  % We devise a coreset selection method based on the idea of gradient matching:
  % the gradients induced by the coreset should match, as closely as possible,
  % those induced by the original training dataset.
  % Inspired by the neural tangent kernel theory, we perform that gradient matching
  % across the model's initialization distribution, allowing us to extract a coreset
  % without having to train the model first.
  % We evaluate the method on a wide range of continual learning scenarios and
  % demonstrate that improves the performance of rehearsal-based methods compared
  % to competing memory management strategies such as reservoir sampling.
\end{abstract}

\section{Introduction}
\label{sec:intro}
The incremental training of machine learning models on non-iid streams of data
is a problem of great practical relevance.
A recommender system may need to be updated with data from new categories of
content or an object detection system may be confronted with new object
categories or changes in lighting conditions or camera position.
In such scenarios, we would like to efficiently update an ML model with new
data without forgetting previously-learned knowledge.

Problems of this type recently received substantial interest from the ML
community under the name of continual learning.
While several specialized formulations of continual learning problems
have been considered, we are interested in the most general one:
We simply assume that we incrementally receive datasets
$\mathcal{D}^{(1)}, \mathcal{D}^{(2)}, \ldots$ without any guarantee of
independence or stationarity.
At time $t$, the goal is to have a model that performs well across all data
seen so far, which we denote as $\mathcal{D}^{(1:t)}$.
The individual datasets are referred to as ``tasks'' in more specialized
formulations, such as class-incremental
(new datasets contain previously-unseen classes)
or domain-incremental (new datasets contain disjoint
parts of the input space) scenarios.
However, we also consider task-free scenarios, where the individual datasets do
not necessarily correspond to conceptually distinct tasks.

With unlimited memory and computational resources, we might just store all
datasets and retrain the model from scratch on $\mathcal{D}^{(1:t)}$.
The challenge lies in getting close to this optimal performance while
processing datasets incrementally, discarding one dataset (or most of it)
before receiving the next.
Naive approaches like simple incremental training on each incoming dataset
lead to so-called \emph{catastrophic forgetting}, the phenomenon of deteriorating
performance on previously-observed data  while training on new data.

Among the most widely-used and effective mechanisms to counteract forgetting is
the use of a rehearsal memory of old data points to be reused while training on
new data.
Rehearsal-based methods are simple, robust, largely model-agnostic, and achieve
state-of-the-art results \citep[e.g.,][]{buzzega2020dark}.
Therefore, the curation of a rehearsal memory is a crucial aspect of continual
learning.
% The ideal memory curation strategy would select a small, representative subset
% of the most ``informative'' data points

In this paper we propose a simple and general coreset selection method for
rehearsal-based continual learning.
Our method is based on the idea of gradient matching:
The gradients induced by the coreset should match, as closely as possible,
those induced by the original dataset.
Inspired by the neural tangent kernel theory, we perform this
gradient matching across the model's initialization distribution.
This allows us to extract coresets without having to train the model and makes the
method independent of the model state and thereby compatible with most, if not
all, rehearsal-based CL strategies.
We demonstrate empirically that gradient-matching coresets improve the performance
of rehearsal-based methods such as GDumb \citep{prabhu2020gdumb} and Experience
Replay \citep{chaudhry2019tiny} compared to other memory management strategies
such as reservoir sampling or sliding window heuristics.

The paper is structured as follows.
Our gradient-matching coreset method (GMC) is introduced in
Section~\ref{sec:gmc}.
Section~\ref{sec:theory} discusses some theoretical considerations, in particularly
the connection to the neural tangent kernel and an interpretation of GMC as
\emph{kernel mean matching}.
Related work is discussed in Section~\ref{sec:related-work}.
In Section~\ref{sec:experiments} we evaluate the method experimentally on a
diverse set of continual learning problems.
We also provide a detailed sensitivity analysis with respect
to the method's hyperparameters.
Finally, in Section~\ref{sec:conclusions} we conclude and suggest some future
research directions.

\section{Gradient-Matching Coresets}
\label{sec:gmc}
We consider a supervised learning problem of predicting targets $y\in \mathcal{Y}$
from inputs $x\in \mathcal{X}$.
We use a model $f(\cdot\,; \theta)\colon \mathcal{X}\to\mathcal{Y}$ parametrized
by $\theta \in \R^D$.
While this could be any parametric model, we are mainly interested in neural
networks.
Given a loss function $l\colon \mathcal{Y}\times\mathcal{Y} \to \R$, any data
point $x, y$ induces a loss $\ell(\theta; x, y) = l(f(x; \theta), y)$.
A training dataset $\mathcal{D}=\{ (x_1, y_1), \dotsc, (x_N, y_N) \}$
implies a training loss
\begin{equation}
  L_{\mathcal{D}}(\theta) = \frac{1}{N} \sum_{i=1}^N \ell(\theta; x_i, y_i).
\end{equation}
During training, the model and the dataset interact via repeated evaluations of
the gradient $\nabla L_\mathcal{D}(\theta)$.
If we could select a subset $\mathcal{C} \subset \mathcal{D}$ such
that $\nabla L_\mathcal{C}(\theta) \approx \nabla L_{\mathcal{D}}(\theta)$ for all
``relevant'' values of $\theta$,
%\gz{wouldn't be enough to select such a set for the optimal set of parameters $\Theta^{\ast}$? of course that's unknown}
%\lb{there's no reason why matching at the optimum should be sufficient}
we could hope to train the model on $\mathcal{C}$
instead of $\mathcal{D}$ with minimal loss in performance.
Inspired by the neural tangent kernel theory \citep{jacot2018neural,arora2019on},
we posit
that the effect of the dataset on the model is, to a certain extent,
characterized by the gradient
across the initialization distribution, which we denote by $p(\theta)$.
Therefore, we define the gradient matching error as
\begin{equation}
  \label{eq:gradient-matching-error}
  \E_{p(\theta)}\left[ \left\Vert \nabla L_\mathcal{C}(\theta) - \nabla L_\mathcal{D}(\theta) \right\Vert^2 \right],
\end{equation}
and seek to select a subset which minimizes this error.

We allow for a \emph{weighted} subset and formalize the selection problem using
a vector $w\in \R^N$
of weights and minimizing
\begin{equation}
  \label{eq:omp-problem-with-expectation}
  \E_{p(\theta)} \left[
  \left\Vert \sum_{i=1}^N w_i \nabla \ell(\theta; x_i, y_i) - \sum_{i=1}^N \nabla \ell(\theta; x_i, y_i) \right\Vert^2
\right]
\end{equation}
subject to a cardinality constraint $\Vert w\Vert_0 \leq n$, where $n\ll N$ is the
desired coreset size and $\Vert w\Vert_0$ denotes the pseudonorm counting
the non-zero entries in $w$.

The remainder of this section will be concerned with the efficient (approximate)
solution of this problem.
We make it tractable by constructing finite-dimensional embeddings of the
per-data point gradient functions.
The resulting cardinality-constrained quadratic problem may be solved
approximately using greedy orthogonal matching pursuit \citep{mallat1993matching},
which we briefly review in Section~\ref{sec:omp}.
In Section~\ref{sec:regularization}, we devise a regularization scheme to avoid
over-reliance on individual data points.
The computational and memory cost of the method are discussed in Section~\ref{sec:cost}.
Finally, we adapt the algorithm to the continual setting in Section~\ref{sec:gmc-continual}

\subsection{Gradient Embeddings}
\label{sec:finite-dimensional-embeddings}

As we cannot compute the expectation in Eq.~\eqref{eq:omp-problem-with-expectation} analytically, we approximate it by Monte Carlo integration:
\begin{equation}
    \frac{1}{S} \sum_{s=1}^{S} \left\Vert
      \sum_{i=1}^N w_i \nabla \ell(\theta_s; x_i, y_i) - \sum_{i=1}^N \nabla \ell(\theta_s; x_i, y_i)
    \right\Vert^2 ,
\end{equation}
where $\theta_1, \dotsc, \theta_S \sim p(\theta)$.
We can write this compactly as
\begin{equation}
  \label{eq:omp-problem-finite}
  \min_{w\in \R^N} \Vert Gw - g \Vert_2^2 \quad \text{s.t.}\quad  \Vert w\Vert_0 \leq n ,
\end{equation}
where we define the \emph{gradient embedding} for the $i$-th data point as the concatenation
of the gradients evaluated at all sampling locations,
\begin{equation}
  \label{eq:grad-embedding-before-proj}
  g_i = \left[\begin{array}{ccc} \nabla \ell(\theta_1; x_i, y_i) \\ \vdots \\ \nabla \ell(\theta_S; x_i, y_i) \end{array}\right] \in \R^{DS},
\end{equation}
as well as $G = \left[ \begin{array}{ccc} g_1 & \cdots & g_N \end{array}\right] \in \R^{DS\times N}$
and $g = \sum_{i=1}^N g_i \in \R^{DS}$.
Eq.~\eqref{eq:omp-problem-finite} is a cardinality-constrained quadratic problem, which is known to be NP-hard,
but approximate solutions may be obtained with greedy methods.

%\gz{the following two paragraphs seems to break a bit the flow -- I would move them in a subsection just before or just after the regularization one}
%\lb{I agree on the flow, but I also don't like to make the reader believe we are using these humongously big gradients only to clear it up two sections later.}
\paragraph{Dimensionality reduction}
For large $D$ and $N$, storing $G$ becomes infeasible.
To alleviate this issue, we project the individual gradients onto a lower-dimensional
subspace using a projection matrix $P\in\R^{d \times D}$, i.e., we replace
Eq.~\eqref{eq:grad-embedding-before-proj} with
\begin{equation}
  g_i = \left[\begin{array}{ccc} P \nabla \ell(\theta_1; x_i, y_i) \\ \vdots \\ P \nabla \ell(\theta_S; x_i, y_i) \end{array}\right] \in \R^{dS}.
\end{equation}
We use sparse random projection matrices as proposed by \citet{achlioptas2001database},
see Appendix~\ref{apx:method-details} for details.
By construction, these projections satisfy $\E[P^TP]=I$ and thereby preserve
inner products in expectation.
Since our algorithm will rely entirely on inner products between gradient
embeddings, this is a crucial property.

\paragraph{Last-layer variant}
As an additional means of reducing both the required memory and the computational
cost of computing the gradient embeddings, we explored using only the gradients
w.r.t.~the last layer of the neural network model.
These can be obtained at the cost of a forward pass through the network, see Appendix~\ref{apx:method-details}. %CA: might want to provide more details here
The last-layer gradient as a proxy for the full gradient has recently been used
in work on active learning \citep{Ash2020Deep} and in the aforementioned
concurrent work by \citet{killamsetty2021grad}.

% Algorithm~\ref{alg:embedding} provides pseudo-code for the computation of the
% gradient embeddings.

% \begin{algorithm}
%   \caption{Initial gradient embedding}
%   \label{alg:embedding}
% \begin{algorithmic}
%   \State $G = \textsc{zeros}(Sd, N)$
%   \For{$s=1,\dotsc, S$}
%     \State Load $\theta_s$
%     \For{$i=1,\dotsc, N$}
%       \State $G[(s-1)d:sd, i] = P \nabla \ell(\theta_s; x_i, y_i)$
%     \EndFor
%   \EndFor
% \end{algorithmic}
% \end{algorithm}

\subsection{Orthogonal matching pursuit}
\label{sec:omp}

To approximately solve Eq.~\eqref{eq:omp-problem-finite}, we use a well-known greedy algorithm called orthogonal matching pursuit (OMP)
\citep{mallat1993matching}, which we briefly outline here.
In the context of OMP, the columns of $G$ are referred to as dictionary elements
and $g$ is the target.
%CA: we could say something about the quality/properties of the greedy solution obtained?

Assume we currently have a coreset, represented by an index set $I\subset [N]$
and corresponding weights $\gamma \in \R^{\vert I \vert}$.
The coreset yields an approximation $g \approx G_I\gamma$, where $G_I$ is the
restriction of $G$ to the columns contained in the index set $I$.
Matching pursuit greedily adds the element which best matches the residual
$r = g - G_I \gamma$.
That is, we add $k_\ast = \argmax_k g_k^Tr$.

\emph{Orthogonal} matching pursuit optimally readjusts the weights of all coreset
elements after each greedy addition by solving $\min_{\gamma} \Vert G_I \gamma - g\Vert^2$.
We stop the coreset construction when the desired size is reached.
Algorithm~\ref{alg:omp} provides pseudo-code.
In practice, OMP can be implemented more efficiently by incrementally updating
the Cholesky decomposition of $G_I^TG_I$ as described in Appendix~\ref{apx:method-details}.

\begin{algorithm}
  \caption{Orthogonal Matching Pursuit}
  \label{alg:omp}
\begin{algorithmic}
  \Function{OMP}{$G = [g_1, \dotsc,  g_N]$, $g$, $n$}
    \State $I \gets ()$ \Comment{Coreset indices}
    \State $\gamma \gets ()$ \Comment{Coreset weights}
    \While{$\vert I \vert < n$}
      \State $r = g - G_I\gamma$ \Comment{Residual}
      \State $k_\ast = \argmax_{k\notin I} \langle g_k, r\rangle$
      \State $I \gets I \cup (k_\ast)$
      \State $\gamma \gets (G_I^T G_I)^{-1}G_I^T g$
    \EndWhile
    \State \Return $I, \gamma$
  \EndFunction
\end{algorithmic}
\end{algorithm}

Of course, we would like the coreset weights to be nonnegative, since training
on negatively-weighted examples would lead to catastrophic effects.
While there are methods for nonnegative matching pursuit \citep[e.g.,][]{yaghoobi2015fast},
they are more costly and complex to implement.
Negative weights were not an issue in our experiements, especially in the
presence of regularization (see below),
so we do not explictly enforce nonnegativity.
As a safeguard, we manually clip the weights at zero in our implementation.
% CA: Is this the right approach? Instead would it not make sense to remove the data points with a negative weight and update gamma again?

\subsection{Regularization}
\label{sec:regularization}

In our application, we want to avoid over-reliance on individual data points in
the form of outsize weights.
This issue becomes particularly obvious if the desired coreset size approaches
or even exceeds the embedding dimension $dS$,
%\gz{I find a bit confusing this $d$ as the projection size -- overall we could explain everything with $D$ and then discuss the fact that we can reduce D to d, but I guess it's a matter of taste}
in which case any (linearly independent)
subset will be able to match the target.
%(This is reflected in the fact that the matrix $G_I^TG_I$ will become singular.)
It is tempting and straight-forward to add an Euclidan regularization
term to the objective in Eq.~\eqref{eq:omp-problem-finite}.
Indeed, this is done in related work \citep{killamsetty2021grad}.
However, encouraging the weights to be small is, in some sense, misguided
in our application.
Instead, we want to encourage the weights to be close to ``uniform'', i.e., to
put roughly equal weight on all coreset elements.

To that end, we first identify the best uniform solution,
$u_\ast = \argmin_{u \in \R} \Vert G_I (u \mathbf{1}) - g\Vert^2$,
where $\mathbf{1}$ denotes a vector of ones.
Then, we set $\gamma$ by minimizing
\begin{equation}
  \Vert G_I \gamma - g\Vert_2^2 + \lambda \Vert \gamma - u_\ast \mathbf{1}\Vert_2^2
\end{equation}
As an added benefit, we are regularizing towards a \emph{positive} weight for
each coreset element.

To the best of our knowledge, such a regularization scheme has not been
discussed in the literature on matching pursuit algorithms and might be of
independent interest.
We discuss in Appendix~\ref{apx:method-details} how this regularization scheme
can easily be integrated into efficient implementations of OMP.

\subsection{Computational and memory cost}
\label{sec:cost}

\paragraph{Memory}
The memory required for the gradient embedding matrix $G$ is $O(NSd)$.
We use values for $S$ in the range of $3$--$10$ and $d = 1000$, making it
feasible to store $G$ even for large datasets.
We also have to store the random projection matrix $P\in \R^{d
\times D}$ but, owing to its sparsity
(Appendix~\ref{apx:method-details}), this requires only $O(\sqrt{D}d)$ memory.
Finally, the individual parameter samples $\theta_1, \dotsc, \theta_S$ are of total size
$DS$ but could easily be stored on disk and loaded turn by turn when
computing the gradient embeddings.

\paragraph{Compute}
The computational cost of obtaining the gradient embeddings is
dominated by the cost of computing the gradients $\nabla \ell(\theta_s; x_i, y_i)$
for each sample $s=1,\dotsc, S$ and data point $i=1,\dotsc, N$.
This corresponds to $S$ epochs of training.
We experimented with $S$ in the range of $3$--$10$, see
also the sensitivity analysis presented in Section~\ref{sec:experiments}.
Hence, the computational cost of obtaining the gradient embeddings is small
compared to the cost of training the network on the full dataset, which typically
takes dozens or hundreds of epochs.
The last-layer variant further reduces the cost by eliminating the
need for a full backward pass.

The computational cost of OMP is dominated by an $O(n^3)$ dependence on the
desired coreset size \citep{rubinstein2008efficient}, which limits the scalability to very large coreset sizes.
In our experiments, we comfortably ran experiments up to $n=10$k.
In continual learning, it is usually desirable to keep the memory small.

%CA: not sure what the added value is of this paragraph
% \paragraph{Scaling further?}
% A simple strategy to scale to larger coreset sizes would be to split the dataset
% into $K$ chunks and extract a coreset of size $n/K$ from each chunk.
% We did not explore this option further for this work.

\subsection{Continual Version}
\label{sec:gmc-continual}

Our main interest lies in continual learning, which is why we would like to
apply GMC to a (non-iid) sequence of data batches, which need to be processed
sequentially.
For each incoming batch, we compute the corresponding gradient embedding matrix
$G^{(t)} = [g^{(t)}_1, \dotsc, g^{(t)}_{N_t}] \in \R^{dS\times N_t}$.
The goal is to maintain a coreset which, after each new batch, matches the
aggregate gradient of all data points seen so far,
which we denote as
\begin{equation}
  \label{eq:target-vector}
  \textstyle g^{(1:t)} := \sum_{s=1}^t \sum_{i=1}^{N_t} g^{(s)}_i.
\end{equation}

Fortunately, our algorithm can be naturally extended to this setting.
Let $C^{(t-1)}$ denote the gradient embedding matrix of the coreset after processing
tasks $1$ through $t-1$.
Upon receiving $G^{(t)}$, we first update the target vector according to
Eq.~\eqref{eq:target-vector}.
We then run OMP with target $g^{(1:t)}$ and dictionary $G = [C^{(t-1)}, G^{(t)}]$.
The gradient embedding matrix of the resulting coreset is stored for reuse in the
next time step.
Algorithm~\ref{alg:streaming} provides pseudo-code.

Compared to an offline setting where $G^{(1)}, \dotsc, G^{(t)}$ are accessible
simultaneously, we use the exact same target vector $g^{(1:t)}$ but a limited
dictionary $[C^{(t-1)}, G^{(t)}]$ instead of $[G^{(1)}, \dotsc, G^{(t-1)}, G^{(t)}]$.
Since $C^{(t-1)}$ is a coreset representative of $G^{(1)}, \dotsc, G^{(t-1)}$,
we can expect the loss in performance to be small.

Note that the algorithm is free to remove and/or reweight elements from
$C^{(t-1)}$, which allows it to flexibly react to new incoming data.
When encountering data that is similar to data already in the coreset, we can
compensate the change in the target vector by reweighting the existing coreset.
When encountering novel data points, the algorithm will drop some coreset elements
to free up capacity.

We want to emphasize that the algorithm extends so seamlessly to the continual
setting because our gradient embeddings are based on the initialization distribution and
therefore \emph{constant}.
This allows us to track the target vector (Eq.~\ref{eq:target-vector}) exactly
and to reuse the gradient embeddings for the coreset across time steps.

\begin{algorithm}
  \caption{Continual GMC}
  \label{alg:streaming}
\begin{algorithmic}
  \State $g \gets 0$
  \State $C \gets []$
  \While{receiving $\mathcal{D}^{(t)} = \{(x^{(t)}_1, y^{(t)}_1), \dotsc, (x^{(t)}_{N_t}, y^{(t)}_{N_t}\}$}
    \State $G = \textsc{gradient-embedding}(\mathcal{D}^{(t)})$
    \State $g \gets g + \sum_i g^{(t)}_i$\Comment{Update target.}
    \State $G \gets [G, C]$ \Comment{Combined dictionary.}
    \State $I, \gamma = \textsc{omp}(G, g, n)$
    \State $C \gets G_I$
  \EndWhile
\end{algorithmic}
\end{algorithm}

% While we do not consider this further here, one can also envision a
% streaming version where data points arrive individually.
% One simply has to collect these points in a fixed-size buffer; once the buffer
% is full, we proceed as above.

\section{Theoretical Connections}
\label{sec:theory}
In this section we discuss two theoretical connections.
First, we show that GMC is related to kernel mean matching w.r.t.~a kernel
defined via the inner product of gradients at initialization.
Secondly, we discuss the connection of this kernel to the neural tangent
kernel \citep[NTK;][]{jacot2018neural,arora2019on}.

\subsection{GMC as Kernel Mean Matching}

\paragraph{Kernel associated with GMC}
For better readability, we introduce the shorthand $z=(x, y)$ for a labeled data
point.
We start by defining a kernel function between two data points as
\begin{equation}
  \label{eq:exact-kernel}
  k_\text{GMC}(z, z^\prime) := \mathbf{E}_{p(\theta)} [\nabla \ell(\theta; z)^T \nabla \ell(\theta; z^\prime)].
\end{equation}
That is, the inner product of the two data points' gradients, averaged over the
initialization distribution $p(\theta)$.
The implicit feature function $\phi$ underlying this kernel maps a data point $z=(x, y)$
to its associated gradient function, $\phi(z) = \nabla \ell(\,\cdot\, ; z) \in \mathcal{G}$
living in the space $\mathcal{G}$ of functions $\R^D \to \R^D$.
If we equip this space with the inner product
$\langle g, g^\prime\rangle_\mathcal{G} = \E_{p(\theta)}[g(\theta)^T g^\prime(\theta)]$,
we recover the kernel via
$k_\text{GMC}(z, z^\prime) = \langle \phi(z), \phi(z^\prime) \rangle_\mathcal{G}$.
Using this notation, our gradient matching objective from
Eq.~\eqref{eq:omp-problem-with-expectation} can be written as
\begin{equation}
  \label{eq:gradient-matching-kernel-formulation}
  \left\Vert \sum_{i=1}^N w_i \phi(z_i) - \sum_{i=1}^N \phi(z_i) \right\Vert_\mathcal{G}^2,
\end{equation}
which we minimize under a cardinality constraint on $w$.

The object $\tfrac{1}{N}\sum_i \phi(z_i)$ is the (empirical) kernel mean embedding of
the data distribution w.r.t.~the kernel associated with $\phi$.
Kernel mean embeddings \citep{smola2007hilbert} represent a distribution in a
reproducing kernel Hilbert
space and can be used for a variety of tasks, e.g., measuring distances between
distributions or performing two-sample tests.%CA: adding refs?

\paragraph{Kernel mean matching}
One use of kernel mean embeddings, which is related to our application, is
kernel mean matching \citep{huang2006correcting}.
Assume we have a kernel $k(x, x^\prime)$ operating on inputs $x$.
Denote the corresponding feature map as $\psi(x)$ which maps to some
Hilbert space $\mathcal{H}$.
Also assume that you have two data samples $x_1, \dotsc, x_N$ and
$x_1^\prime, \dotsc, x_N^\prime$, generated by $p(x)$ and $p^\prime(x^\prime)$,
respectively.
Kernel mean matching finds weights $w$ which minimize
\begin{equation}
  \label{eq:kernel-mean-matching}
  \left\Vert \frac{1}{N} \sum_{i=1}^N w_i \psi(x_i) -
  \frac{1}{N^\prime} \sum_{i=1}^{N^\prime} \psi(x^\prime_i)
  \right\Vert_\mathcal{H}^2
\end{equation}
subject to the constraint $w_i\geq 0$.
In other words, we %want to
reweight the samples from $p(x)$ such as to better match the
kernel mean embedding of $p^\prime(x)$, akin to importance sampling.
A typical application is to correct for a sample selection bias.
%Assume the distribution used for data collection is described by $p(x)$ and you
%know the labels for $x_1,\dotsc, x_N$.
%For a variety of reasons, $p(x)$ may differ from the ``true'' distribution $p^\prime(x)$,
%from which you only have unlabeled examples.
%We can use kernel mean matching to reweight our labeled samples such as to be more representative
%of the true distribution $p^\prime(x)$. %CA: unclear what the added value is of this paragraph -- I commented it out

\paragraph{Comparison}
Comparing Eqs.~\eqref{eq:gradient-matching-kernel-formulation} and
\eqref{eq:kernel-mean-matching}, GMC can be seen as a form of kernel mean matching
where the source and target distributions are \emph{identical} and we impose
a cardinality constraint.
Note, however, that we use a kernel and associated feature function that operates on labeled
data points $z=(x, y)$, whereas existing applications of kernel mean matching
use kernels operating on inputs only.

\subsection{Relationship to the NTK}

Recall that $\ell(\theta; x, y) = l(f(x; \theta), y)$, where
$f(\cdot; \theta)\colon \mathcal{X}\to \mathcal{Y}$ denotes the forward map
of the neural network and $l\colon \mathcal{Y}\times \mathcal{Y} \to \R$ is
the loss function.
The neural tangent kernel is defined for a regression problem---i.e.,
$\mathcal{Y} = \R$ and $l(\hat{y}, y) = \frac{1}{2} (\hat{y} - y)^2$---
and reads
\begin{equation}
  \label{eq:ntk}
  k_\text{NTK}(x, x^\prime) = \nabla_\theta f(x; \theta)^T \nabla_\theta f(x^\prime; \theta),
\end{equation}
evaluated at a location $\theta \sim p(\theta)$ sampled randomly from the
initialization distribution.

The two gradients featuring in Eq.~\eqref{eq:exact-kernel} and Eq.~\eqref{eq:ntk}
are closely related, since
\begin{equation}
  \nabla \ell(\theta; x, y) = l^\prime(f(x; \theta), y) \, \nabla_\theta f(x; \theta).
\end{equation}
However, there are two key differences between the NTK and the GMC kernel.
First, the NTK operates on inputs $x$, whereas the GMC kernel
operates on input-output pairs $(x, y)$.
Second, while the GMC kernel is averaged over $p(\theta)$, the NTK is
evaluated at a single location sampled randomly from $p(\theta)$.
The theory associated with the neural tangent kernel is concerned with the
so-called infinite-width limit, where the number of units in each layer of the
neural network grows to infinity.
While the kernel defined in Eq.~\eqref{eq:ntk} depends
on the (randomly-sampled) $\theta$ for a finite-width model, it converges to a deterministic object in
the infinite-width limit.

\section{Related work}
\label{sec:related-work}
Continual learning has received significant attention in recent
years.
Several approaches to counteract forgetting have been explored, such as
regularization \citep{kirkpatrick2017overcoming, zenke2017continual},
dynamically evolving architectures
\citep{rusu2016progressive, mallya2018packnet, serra2018overcoming}, or
knowledge distillation \citep{li2017learning, rebuffi2017icarl}.

%The use of a rehearsal memory is arguably the single most effective strategy to
%counteract forgetting. -- GZ: this is too strong
% LB: I think that has been demonstrated pretty conclusively. And it's kind of important as motivation, no?
Among the most widely-used methods are those that rely on a so-called
\emph{rehearsal memory} \citep{robins1995catastrophic} of data points to be
reused while training on new data.
Rehearsal-based methods are simple, robust, and largely agnostic to the
model class.
A simple but effective method in this family is \emph{Experience Replay}
\citep{robins1995catastrophic, chaudhry2019tiny}, % there is an early work for ER -- added here
which trains jointly on the current task data and the memory.
Different memory curation strategies can be used but reservoir sampling
\citep{vitter1985random}, which maintains a uniform random
subsample from a stream of data points,
has been found to work best by \citet{chaudhry2019tiny}.
Other works have relied on sliding window heuristics, e.g., \citet{lopez2017gradient}.
In a notable paper, \citet{prabhu2020gdumb} demonstrated that retraining---from
scratch---on a greedy class-balanced memory alone is competitive with a number
of more complex methods;
a finding which underscores the significance of a rehearsal memory.

A number of works provide solutions for curating a rehearsal memory in a more
informed manner.
We limit the discussion to the methods most relevant to the present work.
\citet{aljundi2019gradient} propose gradient-based sample selection (GSS), which
assembles a set of data points with diverse gradients by minimizing the sum of
their pairwise cosine similarities.
\citet{yoon2021online} seek to maximize a combination of gradient diversity and
gradient similarity to the current task.
\citet{borsos2020coresets} construct coresets for continual learning using a
bilevel optimization approach which also relies on matching pursuit.

Outside of continual learning, \citet{campbell2018bayesian} use a similar
technique to select coresets for
efficient Bayesian inference with Monte-Carlo methods.
\citet{zhao2020dataset} use a related notion of gradient matching
to construct \emph{synthetic} datasets.

Recent independent work by \citet{killamsetty2021grad} is closely
related to our method, but pursues a different goal, namely to reduce the
computational cost of training a model in an offline (non-continual) setting.
They use gradient matching to select an ``active set'' of points, which are used
to train the model for a small number of epochs, after which the selection is
repeated.
Their method is based on gradients evaluated \emph{locally} at the
current location in parameter space.
Since the whole dataset has to be retained for the repeated coreset selectio,
their method is not directly applicable to continual learning.
We experimented with a ``local version'' of our algorithm (see
Appendix~\ref{apx:method-details}), which can be seen
as an adaptation of the method of \citet{killamsetty2021grad} to the continual
setting.

\section{Experiments}
\label{sec:experiments}
\begin{figure*}[t]
	\centering
	\includegraphics[width=0.33\textwidth]{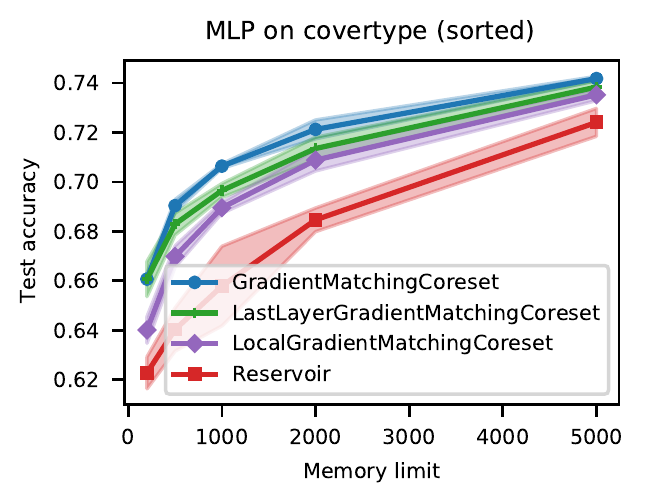}
  \includegraphics[width=0.33\textwidth]{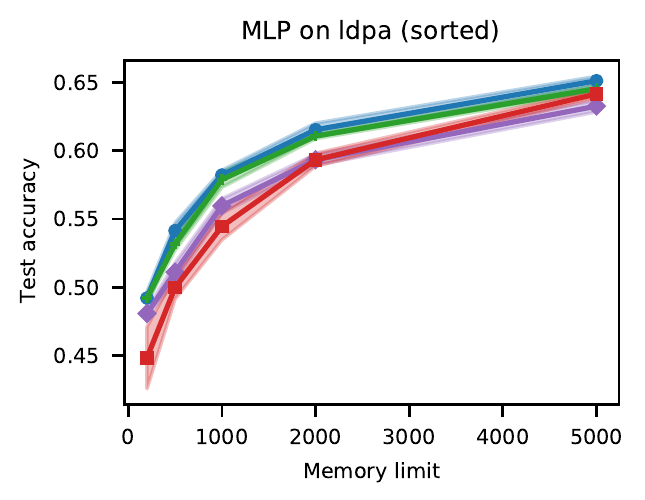}
  \includegraphics[width=0.33\textwidth]{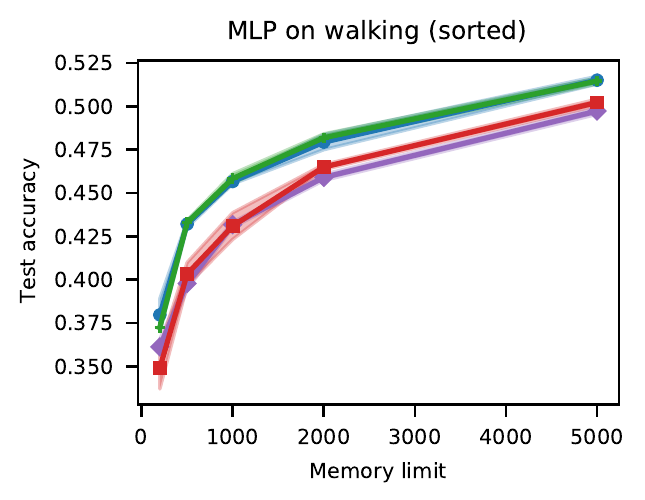}\\
  \includegraphics[width=0.33\textwidth]{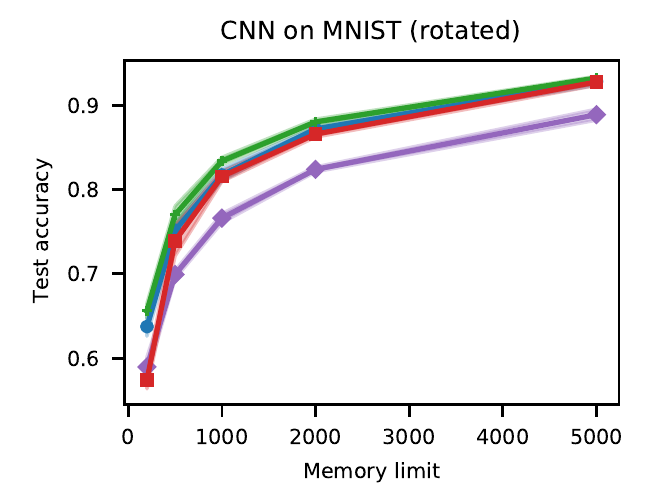}
  \includegraphics[width=0.33\textwidth]{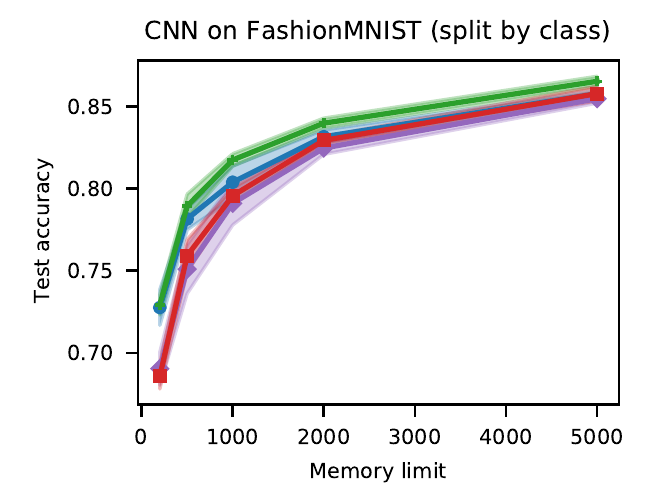}
  \includegraphics[width=0.33\textwidth]{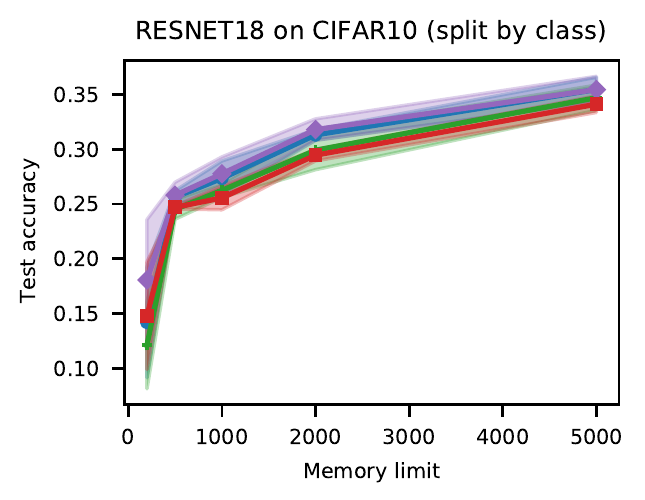}
  \caption{Experimental results for different memory curation strategies in the
  GDumb paradigm. The graphs
  depict the final test accuracy, after seeing all tasks/batches, as a function
  of the memory size. Results are averaged over five random seeds and the
  shaded area spans one standard deviation.}
	\label{fig:results-gdumb}
\end{figure*}

\begin{figure*}
	\centering
	\includegraphics[width=0.33\textwidth]{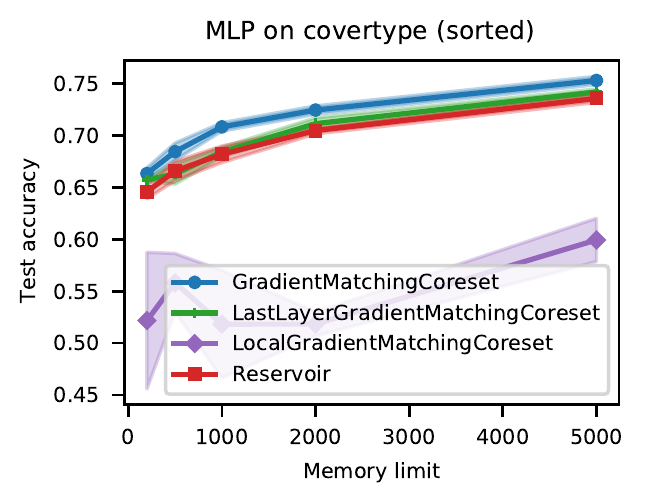}
  \includegraphics[width=0.33\textwidth]{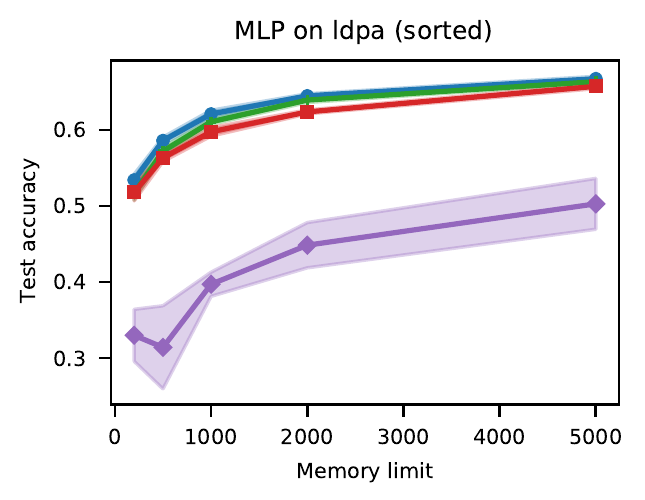}
  \includegraphics[width=0.33\textwidth]{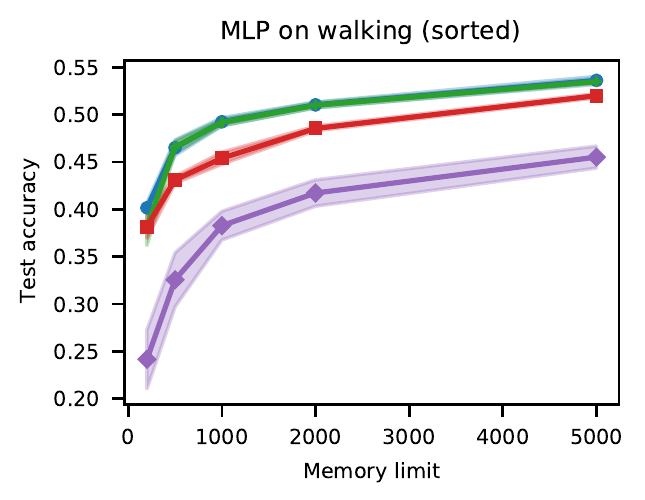}\\
  \includegraphics[width=0.33\textwidth]{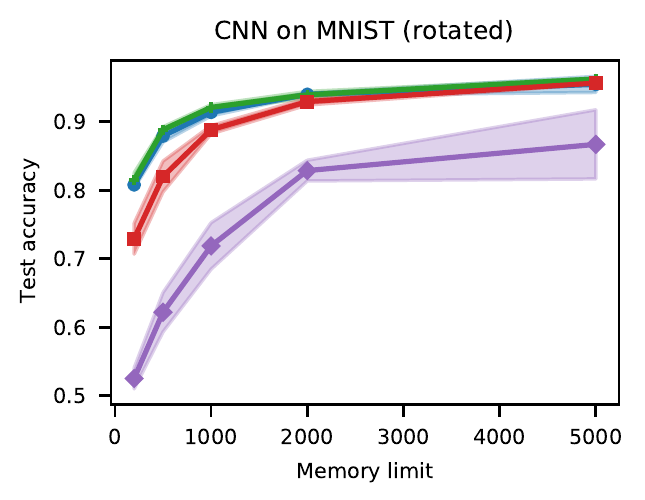}
  \includegraphics[width=0.33\textwidth]{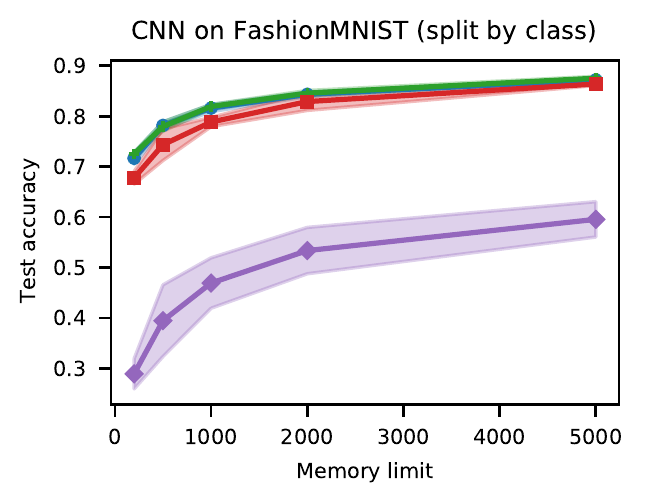}
  \includegraphics[width=0.33\textwidth]{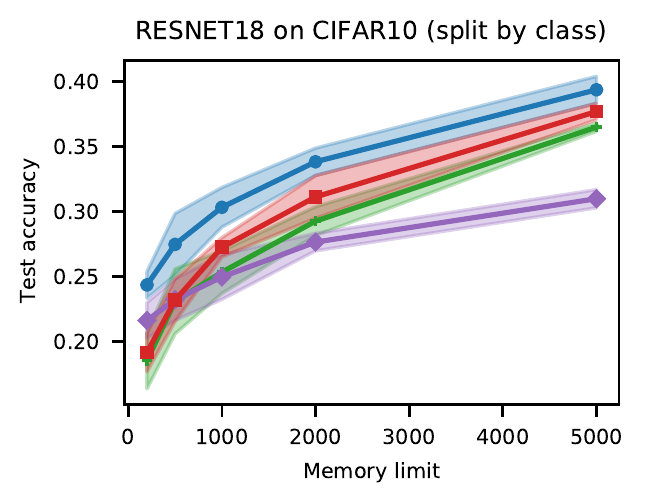}
  \caption{Experimental results for different memory curation strategies in the
  Experience Replay paradigm.
  The graphs
  depict the final accuracy, after seeing all tasks/batches, on
  the full test set as a function of the memory size. Results are averaged over
  five random seeds and the shaded area spans one standard deviation.}
	\label{fig:results-er}
\end{figure*}

We evaluate the performance of GMC against different memory curation strategies
for continual learning.
While GMC may be used with any rehearsal-based CL method, for the
purpose of this comparison we leverage two simple and well-know algorithms: a
variant of GDumb~\citep{prabhu2020gdumb} and Experience Replay~\citep{chaudhry2019tiny}.

\subsection{Experimental Setup}

\paragraph{CL Methods}
Upon receiving a new batch of data, GDumb updates its rehearsal
memory, reinitializes the model and trains it from scratch using only
the data in memory. The rehearsal memory is the \emph{sole} mechanism counteracting
forgetting, and therefore these experiments clearly isolate the effect of the
memory curation strategy.
While \citet{prabhu2020gdumb} use a greedy class-balancing memory, we
use the same approach with other memory curation strategies.

In Experience Replay, we train on the union of the current
task data and the memory.
The model is trained continually without reinitializing.
The memory is updated when training for a task has terminated.

\paragraph{Benchmarks}
We present experiments training an MLP on three tabular datasets,
\textsc{covertype}, \textsc{ldpa}, and \textsc{walking activity},
from OpenML \citep{OpenML2013},
a CNN on \textsc{mnist} \citep{lecun2010mnist} and
\textsc{fashion-mnist} \citep{xiao2017online},
as well as a ResNet-18 \citep{he2016deep} on \textsc{cifar10} \citep{krizhevsky2009learning}.
To demonstrate the versatility of the method, we use a wide variety of continual
learning scenarios.
For the tabular datasets, we generate a challenging task-free scenario
by sorting the data points according to the value of a single feature and splitting
into $10$ equally sized batches.
There is no notion of distinct tasks but a continuous drift in the
input distribution.
%For example, in \textsc{covertype} the feature by which we sort is the
%elevation of the patch of land.
For \textsc{mnist}, we follow \citet{lopez2017gradient} and generate a
domain-incremental scenario by splitting the dataset into $4$ folds and rotating the images by
$0$, $90$, $180$, and $270$ degrees, respectively.
For \textsc{fashion-mnist} and \textsc{cifar-10}, we use the popular class-incremental
scenario, where the dataset is divided into discrete tasks, each consisting of
two classes.
Full experimental details may be found in Appendix~\ref{apx:experimental-details}.

\paragraph{Baselines}
We compare GMC to Reservoir Sampling \citep{vitter1985random},
the greedy class-balancing method used by \citep{prabhu2020gdumb}, a simple sliding
window heuristic,
as well as the gradient-based sample selection (GSS) strategy
of \citet{aljundi2019gradient}.
Lastly, we experimented with a ``local'' variant of GMC, which may be seen as a
continual version of the method of \citet{killamsetty2021grad}.
Details are explained in Appendix~\ref{apx:method-details}, but the basic idea
is to perform gradient matching \emph{locally}, at the current location in
parameter space.
%\gz{probably need to say that you use the model from the previous step for the gradients}.
%\lb{I made this more precise in the appendix.}
All competitors have been tested using memories of 200, 500, 1000, 2000 and
5000 data points.

\paragraph{Hyperparameters}

We used a fixed hyperparameter setting for GMC across all experiments,
identified with a few exploratory experiments.
To compute the gradient embeddings, we used $S=10$ draws from the initialization
distribution and a projection dimension of $d=1000$.
The sensitivity analysis in Fig.~\ref{fig:hyperparameter-sensitivity}
suggests that these are generous values and could likely be reduced without
sacrificing performance. We use a regularization coefficient of $\lambda = 0.5$.

Regarding the optimization procedure, we opted for a single set of hyerparameters
across all experiments.
We used the Adam optimizer \citep{kingma2014adam}
with a constant step size of $3\cdot 10^{-4}$ and other hyperparameters set to
default values.
We employed a weight decay penalty of $10^{-4}$ and used a minibatch size of $100$.
These settings are not tuned to the individual problems and therefore likely
suboptimal, but we prioritized a clean and simple comparison between the memory
curation strategies over state-of-the-art results.

\subsection{Results}
%CA: I wonder whether the performance and forgetting correlate (anti) with the gradient matching error?

Due to space constraints, full results for all memory curation strategies are
reported in Tables~\ref{table:results-gdumb} and \ref{table:results-er} in the
Appendix.
Here, in Figures~\ref{fig:results-gdumb} and \ref{fig:results-er}, we visualize
the comparison between GMC (in its different versions) and
Reservoir Sampling, which proved to be the strongest competitor, in line with
findings by \citet{chaudhry2019tiny}.
We see that GMC achieves consistent improvements
over Reservoir Sampling and the other baselines in all scenarios and across all
tested memory sizes, both in GDumb and Experience Replay.
The size of the effect varies and tends to be larger at small memory sizes.
It is quite substantial in the tabular datasets, e.g., on \textsc{covertype}
with a memory size of $1000$, GMC boosts the test accuracy to roughly 70\% compared to the
65\% achievable with Reservoir Sampling, a relative increase of $7.7\%$.
The increase in performance is more modest in the vision experiments, but still
considerable for small memory sizes, especially in the ER experiments.
In any case, GMC \emph{never} underperforms Reservoir Sampling.

As can be seen in the Tables~\ref{table:results-gdumb} and \ref{table:results-er},
the greedy class-balancing method used in the original GDumb paper
matches Reservoir Sampling in the class-incremental scenarios,
but fails in the task-free and domain-incremental scenarios,
where there is a continual dynamic beyond a shift in class occurences.
The sliding window heuristic performs poorly across the board.

The last-layer variant of GMC matches the performance of the full variant in
most experiments and even performs \emph{better} in a few cases.
However, there is the notable exception of the ResNet experiment, where
it underperforms especially when using Experience Replay.
One might conjecture that the last layer gradient is a poor proxy when using
such deep architectures.

The local version---a naive adaptation of the method of \citet{killamsetty2021grad}
to the continual setting---has mixed performance.
Somewhat surprisingly, it performs decently in some GDumb experiements, but
drastically underperforms when used with ER.
We conjecture that the gradients at a single point
in parameter space are not informative enough to select coresets that are useful
beyond a small number of training epochs.
See also our sensitivity analysis regarding the number of samples $S$.

\subsection{Ablation Studies}

\begin{figure*}[t]
  \centering
  \includegraphics[width=0.33\textwidth]{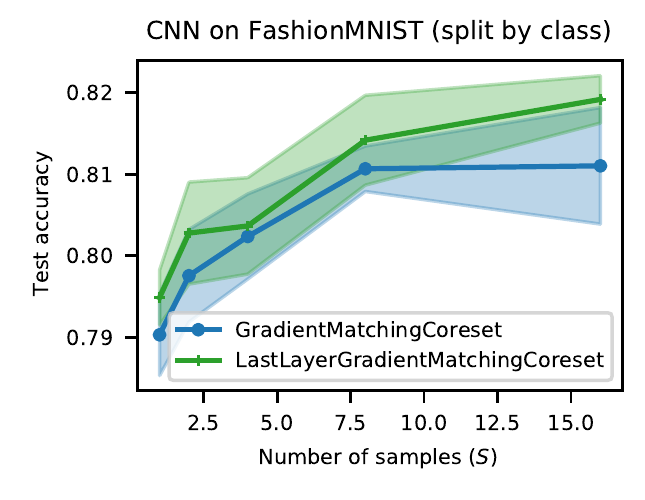}
  \includegraphics[width=0.33\textwidth]{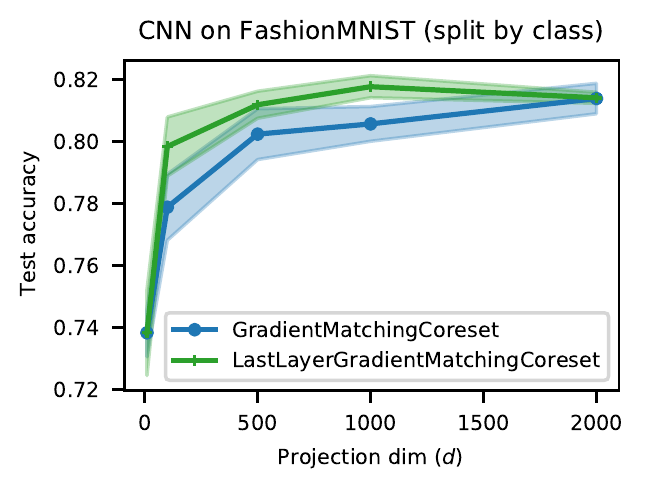}
  \includegraphics[width=0.33\textwidth]{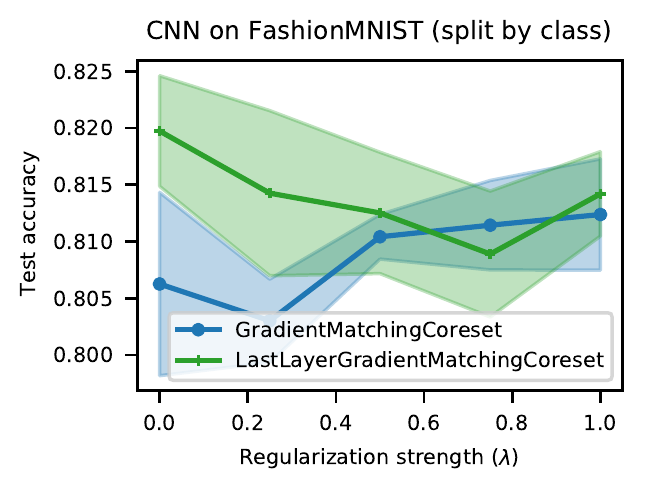}
  \caption{Hyperparameter sensitivity of GMC.}
  \label{fig:hyperparameter-sensitivity}
\end{figure*}

GMC exposes three hyperparameters:
The number of samples from the initialization distribution ($S$),
the projection dimension ($d$) and the regularization strength ($\lambda$).
We ran a (univariate) sensitivity analysis with respect to these hyperparameters
on the \textsc{fashion-mnist} experiment in the GDumb setting at
a memory size of $1000$.
The results are depicted in Figure~\ref{fig:hyperparameter-sensitivity}.
For the gradient embedding parameters ($S$ and $d)$ we see, as
expected, an increasing trend but the effect seems to saturate quite quickly.
%Using $S=1$ leads to a considerable drop in performance.
For the projection dimension, any value larger than $500$ seems to work almost
equally well.
The performance seems to be relatively insensitive to the choice of the
regularization
strength ($\lambda$) in this particular experiment.
However, we can anecdotally report that adding some regularization improves the
robustness of the algorithm.

\begin{figure}
  \centering
  \includegraphics[width=0.4\textwidth]{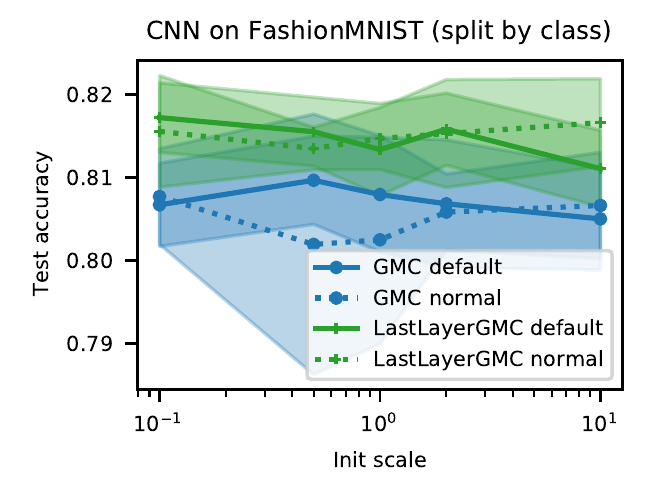}
  \caption{Sensitivity to the type and scale of the initialization distribution.}
  \label{fig:sensitivity-init-scale}
\end{figure}

We also investigated the sensitivity to the choice of the initialization
distribution $p(\theta)$.
In all experiments above, we used the default initialization scheme
implemented in PyTorch (\texttt{kaiming\_uniform}), which is the uniform
version of the scheme proposed by \citet{he2015delving}.
In Figure~\ref{fig:sensitivity-init-scale}, we compare this default scheme to
the corresponding normal (Gaussian) version (\texttt{kaiming\_normal}).
We vary the scale of these distributions by multiplying the
random draws by a scale parameter.
These experiments were run on the \textsc{fashion-mnist}
problem using GDumb with a fixed memory size of $1000$.
The performance of GMC is remarkably insensitive to these variations of the
initialization distribution.

\section{Conclusion}
\label{sec:conclusions}
We introduced the Gradient-Matching Coreset method (GMC).
Our method is simple, robust and scales to coreset sizes of several thousand
data points.
It naturally extends to the continual learning setting, where it can be used
to curate a memory for any rehearsal-based strategy.
We empirically evaluated its performance on a variety of continual learning
scenarios including tabular and image data, balanced and imbalanced datasets,
as well as different types of continual learning scenarios
(class-incremental, domain-incremental, task-free).
GMC is never worse than and often outperforms its strongest competitor,
Reservoir Sampling, across all memory sizes.

There are several avenues for future work.
It would be interesting to explore the use of GMC with
more advanced rehearsal-based methods such as Dark Experience Replay
\citep{buzzega2020dark}.
%
%While we have focused on rehearsal-based methods due to their practical interest,
%GMC can be used out of the box to store subsets of data for other purposes,
%for example for
%knowledge distillation or for methods such as Learning without Forgetting \citep{li2017learning}.
%
While we focussed on classification tasks in this work, GMC is based entirely on
the idea of matching gradients and therefore directly
applicable to other supervised learning tasks (e.g., regression) or unsupervised
learning, e.g., with auto-encoders.
Moreover, it would be interesting to make the coreset size \emph{adaptive} by
replacing the fixed subset size in OMP (Algorithm~\ref{alg:omp}) with
a threshold on the residual error.
This would allow the algorithm to allocate additional memory resources in response
to the ``novelty'' of the incoming data.
Finally, it would be desirable to better understand the algorithm from a
theoretical perspective, e.g., by strengthening the connection to the neural
tangent kernel theory in the infinite width limit.

%\begin{contributions} % will be removed in pdf for initial submission,
                      % so you can already fill it to test with the
                      % ‘accepted’ class option

%\end{contributions}

%\begin{acknowledgements} % will be removed in pdf for initial submission,
                         % so you can already fill it to test with the
                         % ‘accepted’ class option

%\end{acknowledgements}

\cleardoublepage

\bibliography{references.bib}

\cleardoublepage

\appendix

\section{Details on GMC}
\label{apx:method-details}

This section provides a number of details on our gradient-matching coreset method
that have been omitted from the main text.

\subsection{Sparse Random Projections}

As mentioned in Section~\ref{sec:finite-dimensional-embeddings}, we use
dimensionality reduction on the gradient embeddings.
Following \citet{achlioptas2001database}, we use sparse random projections
with a density $\rho \in (0, 1]$.
The elements of $P\in \R^{d\times D}$ are sampled as
\begin{equation}
  P_{ij} = \begin{cases}
      - (\rho d)^{-1/2} & \text{ with probability } \rho/2, \\
      0 & \text{ with probability } 1 - \rho, \\
      (\rho d)^{-1/2} & \text{ with probability } \rho/2.
  \end{cases}
\end{equation}
These projections satisfy $\E[P^TP] = I$ by construction and thereby preserve
inner products in expectation.
Since our algorithm is based entirely on inner products between gradient
embeddings, this is a crucial property.

Following recommendations by \citet{li2006very}, we set $\rho = 1 / \sqrt{D}$
for a good trade-off between memory efficiency and reconstruction quality.
This pushes the memory required to store $P$ down to $O(d\sqrt{D})$.
Note that it would be infeasible to store a dense $d\times D$ matrix for
large model where $D$ is in the tens of millions.

\subsection{Last-Layer Gradients}

The gradients w.r.t.~the last layer's weights can be computed efficiently.
We separate the forward map as $x\mapsto Wh(x; \theta)$, where
$h(x;\theta)$ denotes the penultimate layer representation of data point $x$
and $W\in \R^{C\times H}$ is the weight matrix of the output layer.
Then the loss as a function of $W$ takes the form
\begin{equation}
  \ell(W; x, y) = l (W h(x; \theta), y)
\end{equation}
and the gradient w.r.t.~$W$ is
\begin{equation}
  \nabla_W \ell(W; x, y) = \nabla l(W h(x; \theta), y) \, h(x; \theta)^T,
\end{equation}
where $\nabla l$ denotes the gradient of the loss function w.r.t.~its first argument,
the prediction.

Consider the case of cross-entropy loss, where the predictions take the form of
logits, i.e., unnormalized log probabilities, and the label $y\in\R^C$ is encoded as a one-hot vector.
The loss function may be written as
\begin{equation}
  l(s, y) = - \sum_{c=1}^C y_c \log \rho(s)_c,
\end{equation}
where $\rho$ is the softmax function, i.e., $\rho(s)_c = s_c / (\sum_{c^\prime} s_{c^\prime})$.
The gradient is
\begin{equation}
  \nabla l(s, y) = y - \rho(s).
\end{equation}

These gradients can be obtained efficiently at essentially the cost of a forward
pass through the network.
One simply has to intercept the penultimate layer representations $h(x;\theta)$
during the forward pass.

\subsection{Cholesky-Based OMP}

In each iteration of OMP, we have to invert the matrix $G_I^TG_I$.
This can be done efficiently by incrementally updating the
Cholesky decomposition of this matrix \citep{rubinstein2008efficient}.
When adding an element to the coreset, $I = \tilde{I} \cup \{k\}$, we can split
the matrix like
\begin{equation}
  G_I^TG_I = \left[\begin{array}{cc}
    G_{\tilde{I}}^TG_{\tilde{I}} & G_{\tilde{I}}^Tg_k \\
    g_k^T G_{\tilde{I}} & \Vert g_k\Vert^2
  \end{array}\right].
\end{equation}
For better readability, we abstract this to
\begin{equation}
  A = \left[\begin{array}{cc}
    \tilde{A} & v \\
    v^T & c
  \end{array}\right], \quad \tilde{A} \in \R^{n\times n}, \, v\in \R^n, \, c\in\R.
\end{equation}
If we know the Cholesky decomposition $\tilde{A} = \tilde{L}\tilde{L}^T$, it is
easy to verify that the Cholesky decomposition of $A$ is given by $A=LL^T$, where
\begin{equation}
  L = \left[ \begin{array}{cc}
    \tilde{L} & 0 \\
    w^T & \sqrt{c - \Vert w\Vert^2}
  \end{array}\right], \quad
  w = \tilde{L}^{-1}v.
\end{equation}
To obtain $w$, we have to solve a linear system with a triangular
matrix, which can be done in $O(n^2)$ time.
Subsequently, $A^{-1}b$ can be solved as $(L^T)^{-1}(L^{-1}b)$ with two more
triangular solves.
Overall, this makes the effort for inverting $G_I^TG_I$ quadratic instead of
cubic.

\paragraph{Regularization}
When adding a regularization term to the matching problem as described in
Section~\ref{sec:regularization}, we will have to invert the matrix
$G_I^TG_I + \lambda I$.
It is straight-forward to extend the Cholesky-based implementation to this case.
We now have a matrix of the form
\begin{equation}
  A + \lambda I = \left[\begin{array}{cc}
    \tilde{A} + \lambda I & v \\
    v^T & c + \lambda
  \end{array}\right].
\end{equation}
Assuming we know the Cholesky decomposition $\tilde{A} + \lambda I = \tilde{L}\tilde{L}^T$,
we have $A + \lambda I = LL^T$, where
\begin{equation}
  L = \left[ \begin{array}{cc}
    \tilde{L} & 0 \\
    w^T & \sqrt{c + \lambda - \Vert w\Vert^2}
  \end{array}\right], \quad
  w = \tilde{L}^{-1}v.
\end{equation}

\subsection{Local variant of GMC}

We also experimented with a ``local'' variant of GMC, which we regard as a
naive adaptation of the method of \citet{killamsetty2021grad}
to the continual learning setting.
After finishing training for a task/batch, we compute gradient embeddings locally,
using only the current iterate and run GMC with these embeddings.
Since this corresponds to $S=1$, we use a larger projection dimension $d$ to get
the same gradient embedding dimension as before.

Since the gradient embedding function now \emph{changes} over time, this requires
changes in Continual GMC (Alg.~\ref{alg:streaming}).
Instead of storing the gradient embedding matrix of the coreset for use in the
next iteration, we need to recompute the embeddings in each iteration.
Also, we can not update the target vector (Eq.~\eqref{eq:target-vector}) and
instead approximate it by summing the gradients in the current coreset and
the new batch of data.
Otherwise, the algorithm proceeds as before using these local gradient
embeddings.

\section{Experimental details}
\label{apx:experimental-details}

\subsection{Continual learning scenarios}

As mentioned in the main text, the class-incremental scenario on \textsc{FashionMNIST}
and \textsc{Cifar-10} is obtained by splitting the $10$ classes of the dataset
into $5$ tasks, consisting of classes $\{0, 1\}, \{2, 3\}, \dotsc, \{8, 9\}$.
The algorithm is evaluated on the entire test set.

For the rotated scenario on \textsc{mnist}, we split the dataset randomly into
four folds, then apply rotations of $0$, $90$, $180$ and $270$ degrees,
respectively.
This is done to both the training and the test set and the algorithm is evaluated
on the entire test set.

\begin{figure}
  \centering
  \includegraphics[width=.49\textwidth]{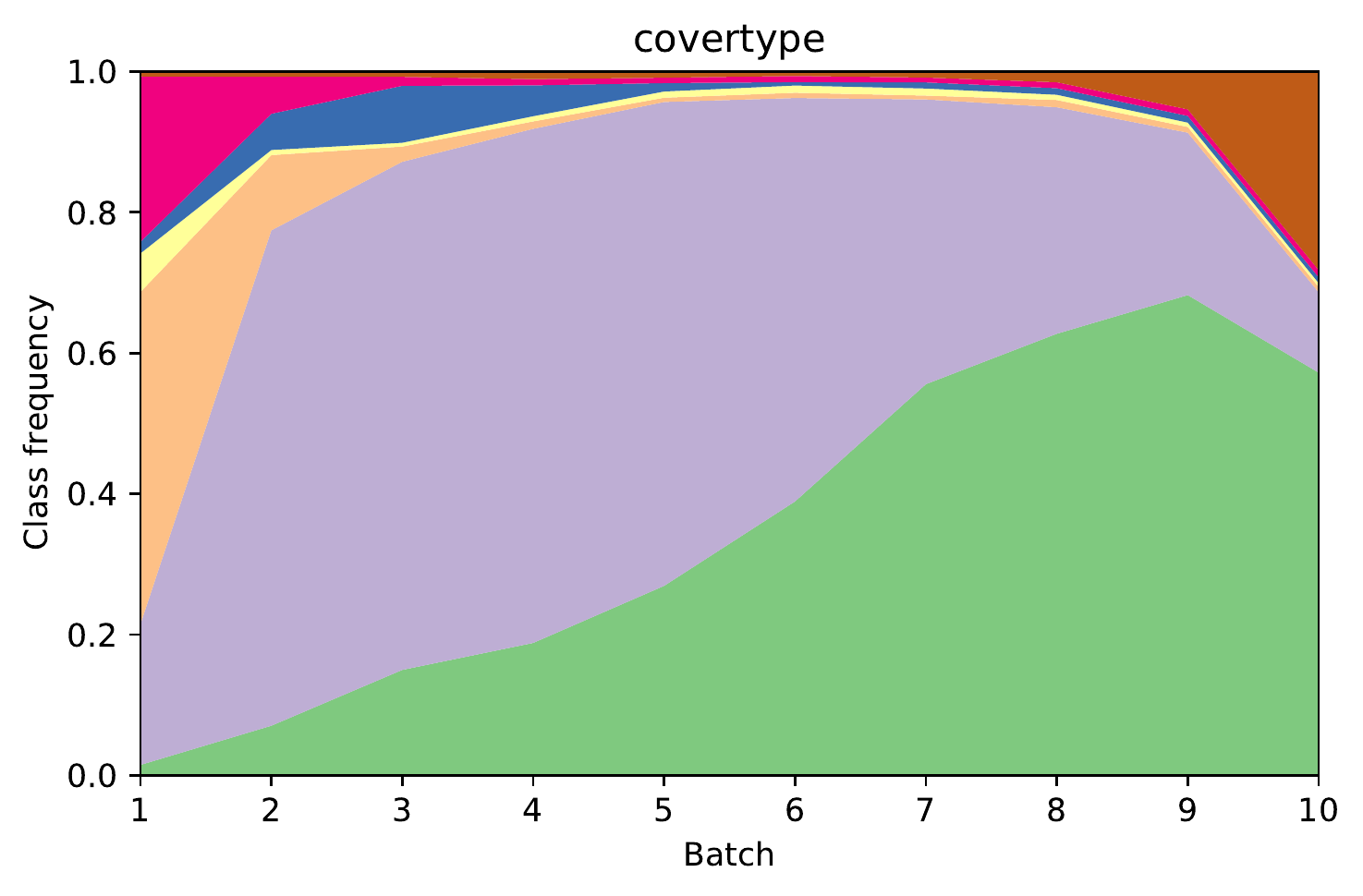}
  \caption{Class frequencies in the different batches of the task-free ``sorted''
  scenario using the Covertype dataset.
  Each color corresponds to on of the seven classes included in the dataset.}
  \label{fig:class-frequencies}
\end{figure}

The task-free scenario for the tabular datasets is generated by sorting the
data points according to the value of a single feature.
We arbitrarily chose the first feature.
The resulting sequence is split into $10$ batches of approximately equal size.
Since the sequence changes smoothly, there is no notion of distinct tasks.
Nevertheless, the sorting generates a non-trivial pattern in the relative frequencies
of the classes across the $10$ batches, see Fig.~\ref{fig:class-frequencies}.
The algorithm is evaluated on the entire test set.

For all datasets, we use the provided train/test split.

\subsection{Model architectures}

All experiments have been implemented using PyTorch \citep{paszke2019pytorch}.
The initialization distribution is the standard scheme implemented in PyTorch,
called \texttt{kaiming\_uniform}, due to \citet{he2015delving}.

\paragraph{MLP}
The MLP architecture consists of two fully-connected hidden layers with 128
units each and ReLU activation, followed by a fully-connected output layer.

\paragraph{CNN}
The CNN architecture consists of three convolutional layers with a receptive
field of $5\times 5$ pixels and $32$, $32$ and
$64$ filters, respectively.
Each convolutional layer is follwed by max pooling over $2\times 2$ windows and
ReLU activation.
This is followed by a fully-connected layer of $256$ units with ReLU activation
and a fully-connected output layer.

\paragraph{ResNet-18} as described by \citet{he2016deep} and implemented in the
\texttt{torchvision} Python package.

\subsection{Optimization hyperparameters}

All models are trained using the Adam optimizer with a step size of $3\cdot 10^{-4}$,
a weight decay of $10^{-4}$ and default choices for the other hyperparameters.
We use a minibatch size of $100$.
We train each model for a fixed number of $200$ epochs, irrespective of the
memory size and algorithm.

\section{Additional results}
\label{apx:additional-results}

\subsection{Full results}

Tables~\ref{table:results-gdumb} and \ref{table:results-er} display the full
results for all tested memory curation strategies using GDumb and ER,
respectively.
Missing entries for the GSS method correspond to training runs that time out.

\begin{table*}
  \centering
  \footnotesize
  \caption{Results in the GDumb setting. Final test
  accuracy [\%] after seeing all tasks for different memory sizes.}
  \label{table:results-gdumb}
  Covertype\\
  \begin{tabular}{llllll}
\toprule
{} &              200  &              500  &              1000 &              2000 &              5000 \\
\midrule
GSS                              &  $30.74 \pm 1.59$ &  $30.79 \pm 2.63$ &  $34.38 \pm 1.86$ &  $39.38 \pm 1.16$ &                 - \\
GradientMatchingCoreset          &  $66.07 \pm 0.22$ &  $69.04 \pm 0.22$ &  $70.64 \pm 0.08$ &  $72.13 \pm 0.34$ &  $74.18 \pm 0.08$ \\
GreedyClassBalancing             &   $4.99 \pm 1.06$ &   $5.84 \pm 1.01$ &   $7.42 \pm 0.24$ &  $12.04 \pm 0.46$ &  $19.75 \pm 0.16$ \\
LastLayerGradientMatchingCoreset &  $66.07 \pm 0.64$ &  $68.29 \pm 0.40$ &  $69.65 \pm 0.25$ &  $71.35 \pm 0.41$ &  $73.84 \pm 0.23$ \\
LocalGradientMatchingCoreset     &  $64.01 \pm 0.49$ &  $66.98 \pm 0.38$ &  $68.95 \pm 0.20$ &  $70.88 \pm 0.39$ &  $73.52 \pm 0.21$ \\
Reservoir                        &  $62.27 \pm 0.58$ &  $64.03 \pm 0.78$ &  $65.78 \pm 1.44$ &  $68.46 \pm 0.43$ &  $72.41 \pm 0.51$ \\
SlidingWindow                    &  $18.23 \pm 0.41$ &  $20.97 \pm 1.03$ &  $21.18 \pm 1.85$ &  $29.98 \pm 0.94$ &  $57.60 \pm 1.81$ \\
\bottomrule
\end{tabular}
\\[8pt]

  LDPA\\
  \begin{tabular}{llllll}
\toprule
{} &              200  &              500  &              1000 &              2000 &              5000 \\
\midrule
GSS                              &  $23.09 \pm 3.91$ &  $21.45 \pm 2.42$ &  $21.91 \pm 2.07$ &  $14.16 \pm 0.44$ &  $16.61 \pm 0.49$ \\
GradientMatchingCoreset          &  $49.21 \pm 0.43$ &  $54.15 \pm 0.52$ &  $58.22 \pm 0.32$ &  $61.55 \pm 0.43$ &  $65.14 \pm 0.29$ \\
GreedyClassBalancing             &  $14.99 \pm 1.59$ &  $10.82 \pm 1.97$ &   $4.17 \pm 0.37$ &   $5.44 \pm 0.14$ &  $10.29 \pm 0.28$ \\
LastLayerGradientMatchingCoreset &  $49.21 \pm 0.30$ &  $53.17 \pm 0.40$ &  $57.90 \pm 0.52$ &  $61.08 \pm 0.17$ &  $64.56 \pm 0.20$ \\
LocalGradientMatchingCoreset     &  $48.07 \pm 0.28$ &  $51.11 \pm 0.61$ &  $55.96 \pm 0.48$ &  $59.35 \pm 0.36$ &  $63.28 \pm 0.38$ \\
Reservoir                        &  $44.83 \pm 2.04$ &  $49.98 \pm 0.76$ &  $54.46 \pm 0.89$ &  $59.34 \pm 0.42$ &  $64.17 \pm 0.29$ \\
SlidingWindow                    &  $16.96 \pm 0.77$ &  $15.76 \pm 0.52$ &  $27.71 \pm 1.29$ &  $32.63 \pm 1.10$ &  $41.89 \pm 1.91$ \\
\bottomrule
\end{tabular}
\\[8pt]

  Walking Activity\\
  \begin{tabular}{llllll}
\toprule
{} &              200  &              500  &              1000 &              2000 &              5000 \\
\midrule
GSS                              &  $14.13 \pm 1.29$ &  $18.13 \pm 1.86$ &  $20.35 \pm 2.39$ &  $15.51 \pm 1.85$ &  $15.75 \pm 2.33$ \\
GradientMatchingCoreset          &  $37.97 \pm 0.83$ &  $43.21 \pm 0.18$ &  $45.67 \pm 0.12$ &  $47.94 \pm 0.39$ &  $51.51 \pm 0.20$ \\
GreedyClassBalancing             &  $13.46 \pm 0.42$ &  $14.27 \pm 1.73$ &  $19.93 \pm 1.59$ &  $19.48 \pm 0.34$ &  $21.22 \pm 0.60$ \\
LastLayerGradientMatchingCoreset &  $37.25 \pm 0.23$ &  $43.29 \pm 0.12$ &  $45.84 \pm 0.24$ &  $48.19 \pm 0.20$ &  $51.46 \pm 0.15$ \\
LocalGradientMatchingCoreset     &  $36.13 \pm 0.39$ &  $39.78 \pm 0.31$ &  $43.17 \pm 0.13$ &  $45.90 \pm 0.16$ &  $49.73 \pm 0.17$ \\
Reservoir                        &  $34.90 \pm 1.08$ &  $40.33 \pm 0.59$ &  $43.09 \pm 0.69$ &  $46.50 \pm 0.13$ &  $50.22 \pm 0.17$ \\
SlidingWindow                    &  $14.73 \pm 0.00$ &  $14.73 \pm 0.00$ &  $17.40 \pm 0.27$ &  $17.37 \pm 0.27$ &  $16.82 \pm 0.13$ \\
\bottomrule
\end{tabular}
\\[8pt]

  MNIST\\
  \begin{tabular}{llllll}
\toprule
{} &              200  &              500  &              1000 &              2000 &              5000 \\
\midrule
GradientMatchingCoreset          &  $63.75 \pm 1.03$ &  $75.16 \pm 0.73$ &  $81.72 \pm 0.63$ &  $87.18 \pm 0.25$ &  $92.82 \pm 0.41$ \\
GreedyClassBalancing             &  $35.09 \pm 1.53$ &  $37.08 \pm 0.65$ &  $38.34 \pm 1.29$ &  $41.42 \pm 0.93$ &  $63.38 \pm 0.67$ \\
LastLayerGradientMatchingCoreset &  $65.63 \pm 0.81$ &  $77.01 \pm 0.93$ &  $83.37 \pm 0.40$ &  $88.01 \pm 0.30$ &  $93.22 \pm 0.18$ \\
LocalGradientMatchingCoreset     &  $58.97 \pm 1.06$ &  $69.92 \pm 0.43$ &  $76.61 \pm 0.49$ &  $82.38 \pm 0.26$ &  $88.87 \pm 0.58$ \\
Reservoir                        &  $57.43 \pm 0.96$ &  $73.92 \pm 1.62$ &  $81.54 \pm 0.53$ &  $86.56 \pm 0.32$ &  $92.74 \pm 0.35$ \\
SlidingWindow                    &  $34.57 \pm 1.48$ &  $36.36 \pm 0.27$ &  $38.34 \pm 1.21$ &  $39.38 \pm 1.00$ &  $42.56 \pm 1.26$ \\
\bottomrule
\end{tabular}
\\[8pt]

  FashionMNIST\\
  \begin{tabular}{llllll}
\toprule
{} &              200  &              500  &              1000 &              2000 &              5000 \\
\midrule
GradientMatchingCoreset          &  $72.76 \pm 0.97$ &  $78.15 \pm 0.59$ &  $80.39 \pm 0.89$ &  $83.16 \pm 0.37$ &  $85.77 \pm 0.18$ \\
GreedyClassBalancing             &  $69.94 \pm 0.66$ &  $75.63 \pm 0.85$ &  $79.74 \pm 0.67$ &  $82.79 \pm 0.59$ &  $85.97 \pm 0.36$ \\
LastLayerGradientMatchingCoreset &  $72.93 \pm 0.55$ &  $78.93 \pm 0.65$ &  $81.74 \pm 0.34$ &  $83.98 \pm 0.27$ &  $86.53 \pm 0.27$ \\
LocalGradientMatchingCoreset     &  $69.04 \pm 0.90$ &  $75.10 \pm 1.33$ &  $79.10 \pm 1.18$ &  $82.46 \pm 0.34$ &  $85.47 \pm 0.24$ \\
Reservoir                        &  $68.59 \pm 0.69$ &  $75.92 \pm 0.88$ &  $79.57 \pm 0.42$ &  $82.93 \pm 0.21$ &  $85.78 \pm 0.39$ \\
SlidingWindow                    &  $19.85 \pm 0.03$ &  $19.91 \pm 0.01$ &  $19.91 \pm 0.01$ &  $19.92 \pm 0.04$ &  $19.96 \pm 0.01$ \\
\bottomrule
\end{tabular}
\\[8pt]

  CIFAR-10\\
  \begin{tabular}{llllll}
\toprule
{} &              200  &              500  &              1000 &              2000 &              5000 \\
\midrule
GradientMatchingCoreset          &  $14.19 \pm 4.53$ &  $25.62 \pm 0.53$ &  $27.34 \pm 1.42$ &  $31.31 \pm 0.43$ &  $35.46 \pm 1.00$ \\
GreedyClassBalancing             &  $16.44 \pm 4.32$ &  $24.38 \pm 0.73$ &  $26.01 \pm 0.52$ &  $29.54 \pm 0.66$ &  $34.43 \pm 0.59$ \\
LastLayerGradientMatchingCoreset &  $12.12 \pm 3.58$ &  $24.52 \pm 0.80$ &  $26.25 \pm 0.58$ &  $29.92 \pm 1.59$ &  $34.71 \pm 1.03$ \\
LocalGradientMatchingCoreset     &  $18.06 \pm 4.94$ &  $25.81 \pm 1.06$ &  $27.76 \pm 1.38$ &  $31.80 \pm 0.88$ &  $35.45 \pm 1.08$ \\
Reservoir                        &  $14.81 \pm 4.40$ &  $24.70 \pm 0.12$ &  $25.56 \pm 0.97$ &  $29.46 \pm 0.47$ &  $34.12 \pm 0.68$ \\
SlidingWindow                    &  $11.28 \pm 1.78$ &  $15.77 \pm 0.25$ &  $15.92 \pm 0.32$ &  $16.42 \pm 0.07$ &  $16.87 \pm 0.10$ \\
\bottomrule
\end{tabular}
\\[8pt]
\end{table*}

\begin{table*}
  \centering
  \footnotesize
  \caption{Experimental results in the experience replay setting. Table shows final test
  accuracy [\%] after seeing all tasks/batches for different memory sizes.}
  \label{table:results-er}
  Covertype\\
  \begin{tabular}{llllll}
\toprule
{} &              200  &              500  &              1000 &              2000 &              5000 \\
\midrule
GSS                              &  $58.99 \pm 1.39$ &  $58.06 \pm 1.57$ &  $57.02 \pm 1.19$ &  $56.97 \pm 2.10$ &                 - \\
GradientMatchingCoreset          &  $66.33 \pm 0.49$ &  $68.44 \pm 0.74$ &  $70.83 \pm 0.37$ &  $72.45 \pm 0.36$ &  $75.30 \pm 0.36$ \\
GreedyClassBalancing             &  $36.00 \pm 3.30$ &  $17.08 \pm 1.46$ &  $17.12 \pm 0.18$ &  $20.17 \pm 0.25$ &  $26.82 \pm 0.33$ \\
LastLayerGradientMatchingCoreset &  $65.72 \pm 0.13$ &  $66.29 \pm 0.87$ &  $68.42 \pm 0.40$ &  $71.14 \pm 0.50$ &  $74.20 \pm 0.26$ \\
LocalGradientMatchingCoreset     &  $52.16 \pm 5.89$ &  $55.81 \pm 2.52$ &  $51.82 \pm 4.64$ &  $51.82 \pm 1.04$ &  $59.93 \pm 1.86$ \\
Reservoir                        &  $64.58 \pm 0.62$ &  $66.58 \pm 0.83$ &  $68.19 \pm 0.72$ &  $70.48 \pm 0.29$ &  $73.57 \pm 0.35$ \\
SlidingWindow                    &  $57.10 \pm 1.63$ &  $57.12 \pm 1.45$ &  $59.10 \pm 0.88$ &  $58.12 \pm 0.86$ &  $49.94 \pm 1.86$ \\
\bottomrule
\end{tabular}
\\[12pt]

  LDPA\\
  \begin{tabular}{llllll}
\toprule
{} &              200  &              500  &              1000 &              2000 &              5000 \\
\midrule
GSS                              &  $43.03 \pm 2.40$ &  $38.65 \pm 2.20$ &  $36.82 \pm 1.02$ &  $39.23 \pm 0.47$ &  $47.56 \pm 1.27$ \\
GradientMatchingCoreset          &  $53.42 \pm 0.68$ &  $58.61 \pm 0.41$ &  $62.10 \pm 0.44$ &  $64.49 \pm 0.24$ &  $66.75 \pm 0.28$ \\
GreedyClassBalancing             &  $37.83 \pm 1.29$ &  $30.16 \pm 2.23$ &  $14.75 \pm 1.24$ &  $14.19 \pm 0.48$ &  $17.78 \pm 0.20$ \\
LastLayerGradientMatchingCoreset &  $51.67 \pm 0.80$ &  $57.22 \pm 0.49$ &  $61.08 \pm 0.30$ &  $63.93 \pm 0.28$ &  $66.36 \pm 0.19$ \\
LocalGradientMatchingCoreset     &  $33.00 \pm 3.05$ &  $31.41 \pm 4.89$ &  $39.71 \pm 1.42$ &  $44.85 \pm 2.66$ &  $50.29 \pm 2.99$ \\
Reservoir                        &  $51.79 \pm 1.00$ &  $56.33 \pm 0.43$ &  $59.76 \pm 0.51$ &  $62.38 \pm 0.23$ &  $65.74 \pm 0.27$ \\
SlidingWindow                    &  $40.96 \pm 1.80$ &  $42.94 \pm 1.61$ &  $43.79 \pm 1.64$ &  $40.90 \pm 1.29$ &  $43.23 \pm 0.40$ \\
\bottomrule
\end{tabular}
\\[12pt]

  Walking Activity\\
  \begin{tabular}{llllll}
\toprule
{} &              200  &              500  &              1000 &              2000 &              5000 \\
\midrule
GSS                              &  $18.81 \pm 0.52$ &  $16.94 \pm 0.74$ &  $17.84 \pm 0.61$ &  $22.54 \pm 0.71$ &  $27.26 \pm 0.49$ \\
GradientMatchingCoreset          &  $40.16 \pm 0.50$ &  $46.50 \pm 0.74$ &  $49.24 \pm 0.39$ &  $51.02 \pm 0.27$ &  $53.60 \pm 0.39$ \\
GreedyClassBalancing             &  $20.69 \pm 1.25$ &  $20.17 \pm 1.21$ &  $22.73 \pm 0.66$ &  $23.82 \pm 0.41$ &  $27.52 \pm 0.43$ \\
LastLayerGradientMatchingCoreset &  $38.17 \pm 1.79$ &  $46.59 \pm 0.57$ &  $49.17 \pm 0.31$ &  $50.99 \pm 0.26$ &  $53.47 \pm 0.17$ \\
LocalGradientMatchingCoreset     &  $24.17 \pm 2.80$ &  $32.56 \pm 2.56$ &  $38.28 \pm 1.36$ &  $41.72 \pm 1.25$ &  $45.51 \pm 1.04$ \\
Reservoir                        &  $38.18 \pm 1.10$ &  $43.11 \pm 0.33$ &  $45.39 \pm 0.57$ &  $48.54 \pm 0.22$ &  $51.99 \pm 0.20$ \\
SlidingWindow                    &  $21.02 \pm 0.42$ &  $21.80 \pm 0.60$ &  $20.92 \pm 0.98$ &  $21.81 \pm 0.37$ &  $22.17 \pm 0.21$ \\
\bottomrule
\end{tabular}
\\[12pt]

  MNIST\\
  \begin{tabular}{llllll}
\toprule
{} &              200  &              500  &              1000 &              2000 &              5000 \\
\midrule
GSS                              &  $63.04 \pm 1.22$ &  $65.02 \pm 0.39$ &  $66.36 \pm 0.76$ &  $67.05 \pm 0.43$ &  $67.42 \pm 0.51$ \\
GradientMatchingCoreset          &  $80.80 \pm 0.41$ &  $87.93 \pm 0.80$ &  $91.39 \pm 0.47$ &  $94.02 \pm 0.12$ &  $95.47 \pm 1.05$ \\
GreedyClassBalancing             &  $63.11 \pm 0.75$ &  $64.77 \pm 0.20$ &  $66.08 \pm 0.77$ &  $66.93 \pm 0.44$ &  $79.38 \pm 0.74$ \\
LastLayerGradientMatchingCoreset &  $81.51 \pm 0.94$ &  $88.84 \pm 0.38$ &  $92.11 \pm 0.35$ &  $93.96 \pm 0.44$ &  $96.27 \pm 0.13$ \\
LocalGradientMatchingCoreset     &  $52.53 \pm 1.40$ &  $62.20 \pm 2.63$ &  $71.88 \pm 3.15$ &  $82.88 \pm 1.35$ &  $86.72 \pm 4.52$ \\
Reservoir                        &  $72.93 \pm 1.96$ &  $82.01 \pm 2.02$ &  $88.85 \pm 0.53$ &  $92.96 \pm 0.45$ &  $95.67 \pm 0.12$ \\
SlidingWindow                    &  $62.13 \pm 0.92$ &  $65.36 \pm 0.52$ &  $66.04 \pm 0.97$ &  $66.82 \pm 0.43$ &  $67.72 \pm 0.54$ \\
\bottomrule
\end{tabular}
\\[12pt]

  FashionMNIST\\
  \begin{tabular}{llllll}
\toprule
{} &              200  &              500  &              1000 &              2000 &              5000 \\
\midrule
GSS                              &                 - &                 - &  $39.00 \pm 0.00$ &  $39.42 \pm 0.01$ &                 - \\
GradientMatchingCoreset          &  $71.67 \pm 1.06$ &  $78.16 \pm 0.66$ &  $81.61 \pm 0.64$ &  $84.33 \pm 0.41$ &  $87.20 \pm 0.26$ \\
GreedyClassBalancing             &  $67.03 \pm 0.65$ &  $74.86 \pm 1.08$ &  $79.35 \pm 0.81$ &  $83.30 \pm 0.37$ &  $86.55 \pm 0.30$ \\
LastLayerGradientMatchingCoreset &  $72.52 \pm 0.26$ &  $77.95 \pm 0.84$ &  $81.92 \pm 0.36$ &  $84.56 \pm 0.48$ &  $87.53 \pm 0.33$ \\
LocalGradientMatchingCoreset     &  $28.93 \pm 2.68$ &  $39.47 \pm 6.36$ &  $46.93 \pm 4.47$ &  $53.36 \pm 4.11$ &  $59.55 \pm 3.13$ \\
Reservoir                        &  $67.78 \pm 1.12$ &  $74.34 \pm 2.80$ &  $78.84 \pm 0.82$ &  $82.89 \pm 1.57$ &  $86.38 \pm 0.27$ \\
SlidingWindow                    &  $35.17 \pm 0.21$ &  $36.82 \pm 0.15$ &  $37.91 \pm 0.14$ &  $38.61 \pm 0.07$ &  $39.00 \pm 0.13$ \\
\bottomrule
\end{tabular}
\\[12pt]

  CIFAR-10\\
  \begin{tabular}{llllll}
\toprule
{} &              200  &              500  &              1000 &              2000 &              5000 \\
\midrule
GradientMatchingCoreset          &  $24.35 \pm 0.92$ &  $27.47 \pm 2.13$ &  $30.33 \pm 1.37$ &  $33.83 \pm 0.94$ &  $39.38 \pm 0.93$ \\
GreedyClassBalancing             &  $19.44 \pm 2.00$ &  $22.71 \pm 0.90$ &  $28.29 \pm 0.89$ &  $30.89 \pm 0.56$ &  $36.90 \pm 0.61$ \\
LastLayerGradientMatchingCoreset &  $18.54 \pm 1.96$ &  $23.09 \pm 2.24$ &  $25.34 \pm 1.47$ &  $29.27 \pm 0.99$ &  $36.50 \pm 0.29$ \\
LocalGradientMatchingCoreset     &  $21.62 \pm 1.23$ &  $23.22 \pm 1.43$ &  $24.96 \pm 1.54$ &  $27.65 \pm 0.60$ &  $30.99 \pm 0.63$ \\
Reservoir                        &  $19.15 \pm 1.32$ &  $23.18 \pm 1.45$ &  $27.24 \pm 0.66$ &  $31.13 \pm 1.47$ &  $37.70 \pm 0.56$ \\
SlidingWindow                    &  $20.73 \pm 1.19$ &  $24.18 \pm 0.78$ &  $26.65 \pm 0.38$ &  $27.91 \pm 0.61$ &  $30.56 \pm 0.26$ \\
\bottomrule
\end{tabular}
\\[12pt]
\end{table*}

\subsection{Continual performance}

\begin{figure*}
  \centering
  \includegraphics[width=0.33\textwidth]{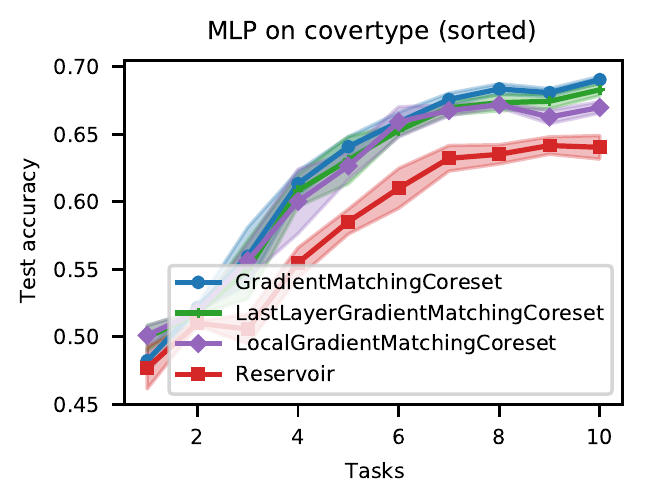}
  \includegraphics[width=0.33\textwidth]{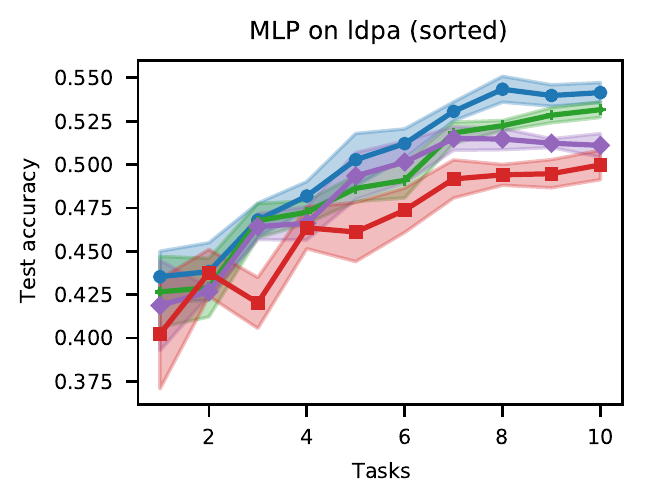}
  \includegraphics[width=0.33\textwidth]{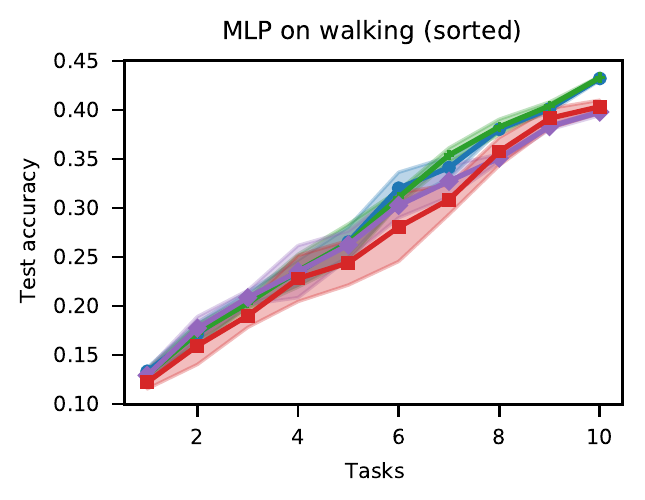}\\
  \includegraphics[width=0.33\textwidth]{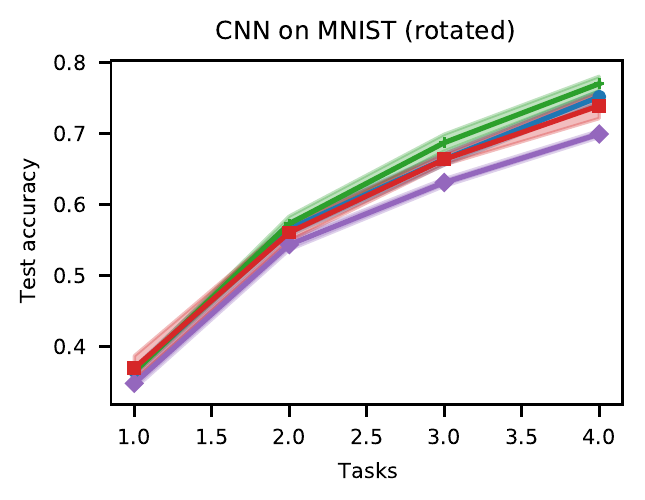}
  \includegraphics[width=0.33\textwidth]{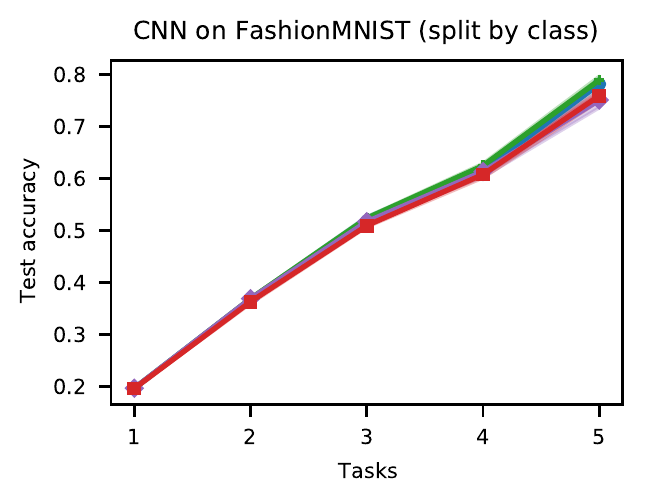}
  \includegraphics[width=0.33\textwidth]{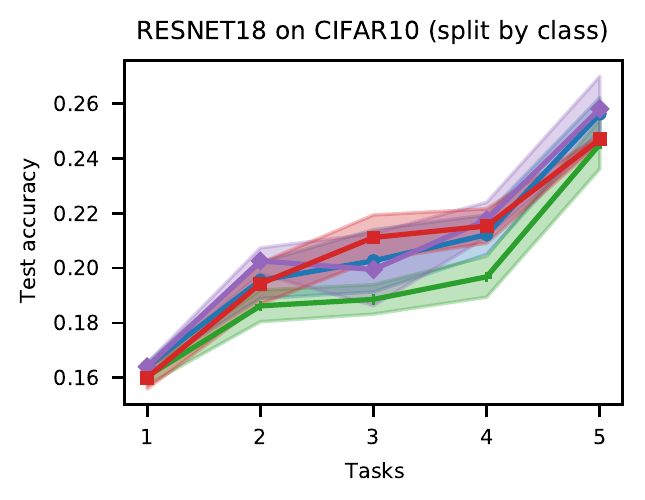}
  \caption{Results on continual learning scenarios using the GDumb
  paradigm with different subsampling/coreset methods at a memory size of $500$.
  The graphs depict the accuracy on
  the full test set after the processing of each task/batch. Results are averaged over
  five random seeds and the shaded area spans one standard deviation.}
  \label{fig:gdumb-results-per-task}
\end{figure*}

The plots in the main text depict the performance after processing all tasks.
To get a more fine-grained few of the continual behavior,
Figures~\ref{fig:gdumb-results-per-task} and \ref{fig:er-results-er-task}
depict the performance after each individual task.
It shows results from the same experiments as in the main text, but zooms in
on a single memory size of $500$.

\begin{figure*}
  \centering
  \includegraphics[width=0.33\textwidth]{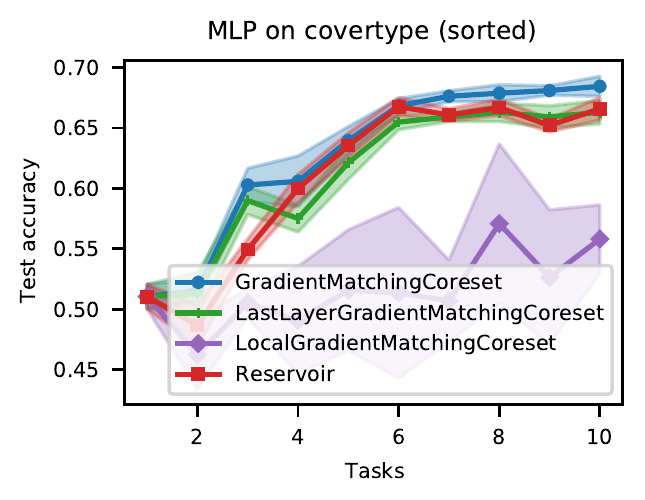}
  \includegraphics[width=0.33\textwidth]{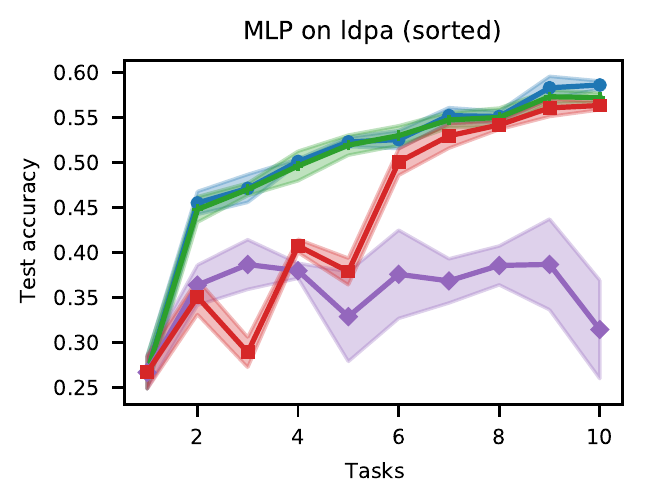}
  \includegraphics[width=0.33\textwidth]{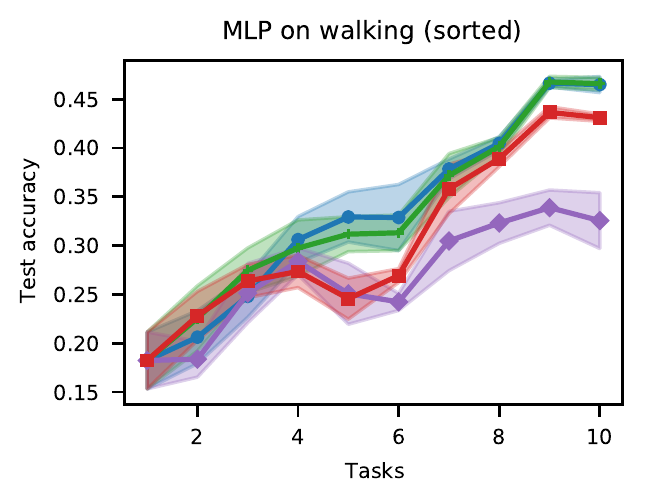}\\
  \includegraphics[width=0.33\textwidth]{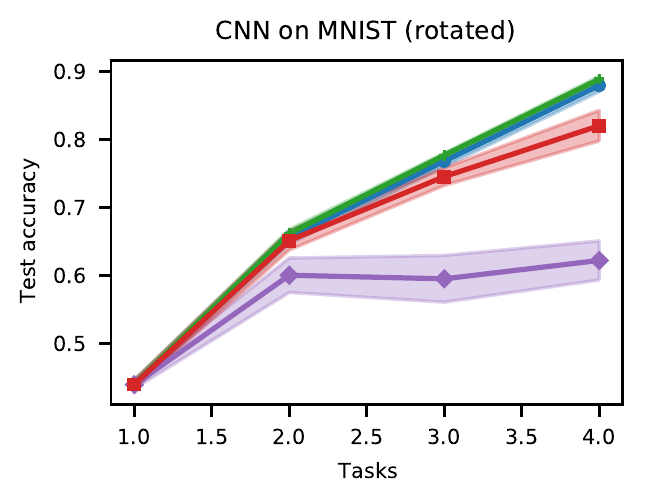}
  \includegraphics[width=0.33\textwidth]{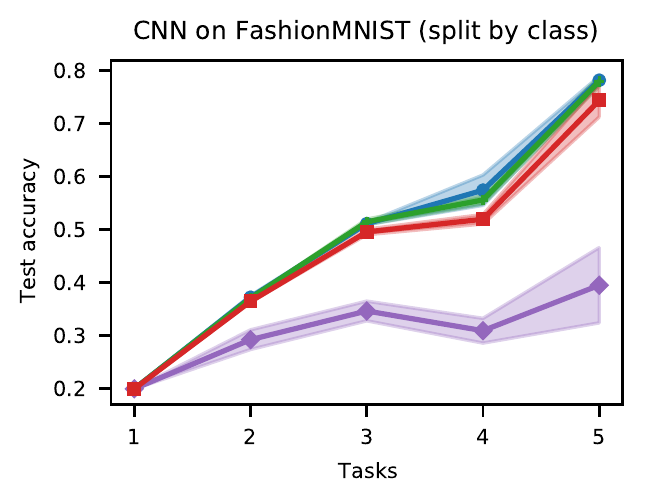}
  \includegraphics[width=0.33\textwidth]{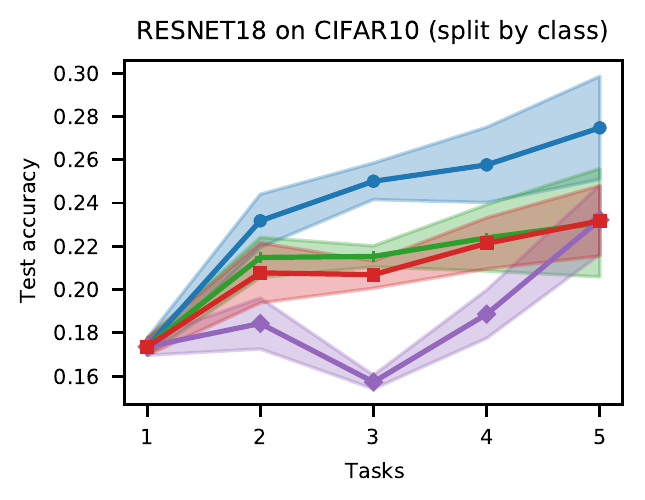}
  \caption{Results on continual learning scenarios using the Experience Replay
  paradigm with different subsampling/coreset methods at a memory size of $500$.
  The graphs depict the accuracy on
  the full test set after the processing of each task/batch. Results are averaged over
  five random seeds and the shaded area spans one standard deviation.}
  \label{fig:er-results-per-task}
\end{figure*}

\subsection{Additional Ablations}

\paragraph{Regularization strategy}

\begin{figure}
  \centering
  \includegraphics{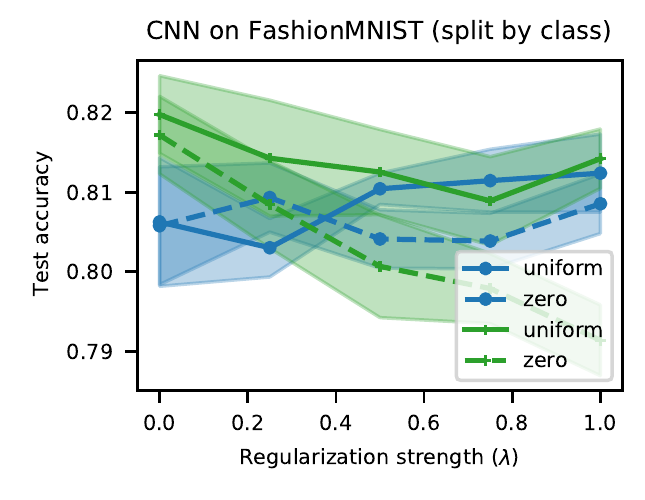}
  \caption{Comparing regularization schemes.}
  \label{fig:ablation-regularization-scheme}
\end{figure}

In Section~\ref{sec:regularization}, we proposed a regularization scheme that
regularizes OMP towards a uniform solution.
In this ablation study, we compare this to a simple zero-centered regularization
term as used by \citet{killamsetty2021grad}.
Results are depicted in Fig.~\ref{fig:ablation-regularization-scheme}.
While the differences are not enormous, the zero-centered regularizer performs
slightly worse.
We also observed that zero-centered regularization led to (a small number) of
coreset elements being assigned a negative weight and therefore being
effectively removed by clipping the weights at zero.

\end{document}